%% file: acl_latex.tex
\PassOptionsToPackage{dvipsnames}{xcolor}
\documentclass[11pt]{article}

\usepackage{acl}

\usepackage{times}
\usepackage{latexsym}

\usepackage[T1]{fontenc}

\usepackage[utf8]{inputenc}

\usepackage{microtype}

\usepackage{inconsolata}

\usepackage{graphicx}

\usepackage{booktabs}

\usepackage[dvipsnames]{xcolor}
\usepackage{enumerate}
\usepackage{pifont}
\usepackage{wrapfig}
\usepackage[utf8]{inputenc} 
\usepackage[T1]{fontenc}    
\usepackage{url}            
\usepackage{amsfonts}       
\usepackage{nicefrac}       
\usepackage{microtype}      
\usepackage{xcolor}         
\usepackage{tikz}

\usepackage{enumitem}
\usepackage{rotating}
\usepackage{epigraph}
\usepackage{adjustbox}

\usepackage{wrapfig}
\usepackage{placeins}  

\usepackage[most]{tcolorbox}  
\tcbset{colback=gray!12, colframe=black}  

\newtcolorbox{obsbox}[1][]{
  enhanced, breakable,
  boxrule=0.8pt, arc=2mm,   
  left=1em, right=1em, top=0.6em, bottom=0.6em,
  colback=gray!12, colframe=black,
  #1
}

\usepackage{amsmath,amssymb,amsfonts}
\usepackage{multirow}
\usepackage{bm}
\usepackage{algorithm}
\usepackage{algorithmic}
\usepackage{colortbl}
\usepackage{array}

\definecolor{lightblue}{RGB}{194, 223, 255}

\usepackage{etoolbox}

\title{ClueTracer: Question-to-Vision Clue Tracing for Training-Free Hallucination Suppression in Multimodal Reasoning}

\author{
  \textbf{Gongli Xi\textsuperscript{1}},
  \textbf{Kun Wang\textsuperscript{2}},
  \textbf{Zeming Gao\textsuperscript{1}},
  \textbf{Huahui Yi\textsuperscript{2}},
\\
  \textbf{Haolang Lu\textsuperscript{1}},
  \textbf{Qianshan Wei\textsuperscript{3}},
  \textbf{Ye Tian\textsuperscript{1}},
  \textbf{Wendong Wang\textsuperscript{1}},
\\
\\
  \textsuperscript{1}Beijing University of Posts and Telecommunications,
  \textsuperscript{2}Nanyang Technological University,
  \\
  \textsuperscript{3}Institute of Automation, Chinese Academy of Sciences
\\
  \small{
  \textbf{Correspondence:} 
  \href{mailto:yetian@bupt.edu.cn}{Ye Tian},
  \href{mailto:wang.kun@ntu.edu.sg}{Kun Wang}
}
}

\begin{document}
\def\method{\textsc{ClueTracer}}
\def\metric{\textsc{ClueRecall}}

\makeatletter
\newcommand{\eg}{\textit{e.g.}\@}
\newcommand{\ie}{\textit{i.e.}\@}
\newcommand{\etal}{\textit{et al.}\@}
\newcommand{\etc}{\textit{etc.}\@}
\makeatother

\maketitle
\input{sections/1_abstract}
\input{sections/2_introduction}
\input{sections/3_related_work}

\input{sections/4_method}
\input{sections/5_experiment}

\input{sections/6_conclusion}

\bibliography{custom}

\appendix

\input{sections/7_appendix}
\label{sec:appendix}

\end{document}

%% file: sections/1_abstract.tex
\begin{abstract}
Large multimodal reasoning models solve challenging visual problems via explicit long-chain inference: they gather visual clues from images and decode clues into textual tokens. 
Yet this capability also increases hallucinations, where the model generates content that is not supported by the input image or the question. To understand this failure mode, we identify \emph{reasoning drift}: during clue gathering, the model over-focuses on question-irrelevant entities, diluting focus on task-relevant cues and gradually decoupling the reasoning trace from visual grounding. As a consequence, many inference-time localization or intervention methods developed for non-reasoning models fail to pinpoint the true clues in reasoning settings. Motivated by these insights, we introduce \metric{}, a metric for assessing visual clue retrieval, and present \method{}, a training-free and architecture-agnostic plugin for hallucination suppression. \method{} starts from the question and traces how key clues propagate along the model’s reasoning pathway (question $\rightarrow$ outputs $\rightarrow$ visual tokens), thereby localizing task-relevant patches while suppressing spurious attention to irrelevant regions. Remarkably, \textbf{without any additional training}, \method{} improves all \textbf{reasoning} architectures (including \texttt{R1-OneVision}, \texttt{Ocean-R1}, \texttt{MM-Eureka}, \etc) by $\mathbf{1.21\times}$ on reasoning benchmarks. Our code is available at \href{https://github.com/Lrbomchz/emnlp_cluereacer}{this link}.
\end{abstract}

%% file: sections/2_introduction.tex
\section{Introduction}

Recent MLLMs have evolved from image description toward cross-modal reasoning, giving rise to Multimodal Large Reasoning Models (MLRMs) that produce multi-step reasoning traces over visual and textual inputs~\cite{xu2024llava, huang2025vision}. These reasoning abilities are typically enhanced through supervised fine-tuning or reinforcement learning on pretrained multimodal backbones~\cite{zhao2024marco, yao2024mulberry, xu2025redstar, yu2024rlhf, MLRM-r1}.


\begin{figure*}[t]
    \centering
    \includegraphics[width=.9\textwidth]{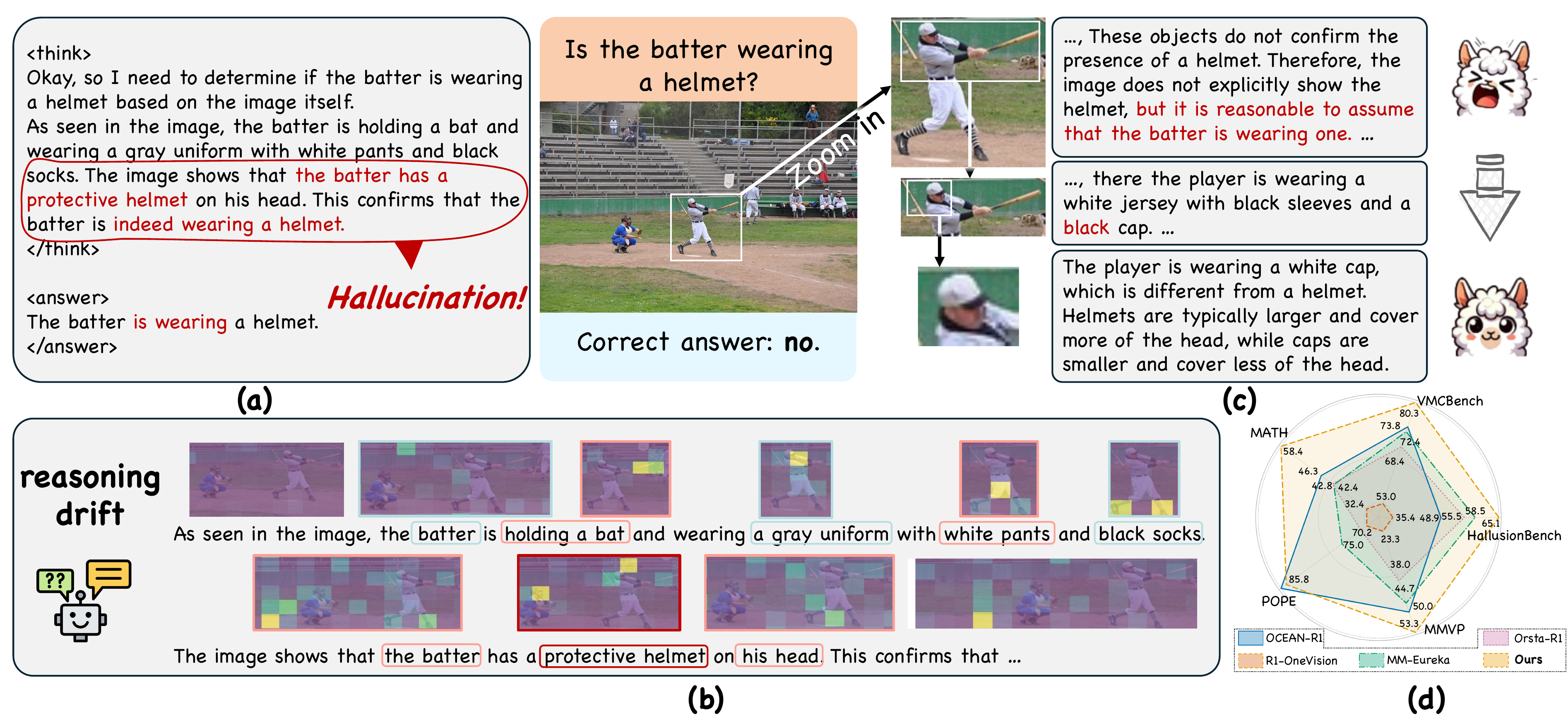}
    \caption{\textbf{(a)} Example outputs from a reasoning model; hallucinated content is highlighted in red. \textbf{(b)} Reasoning drift in a multimodal reasoning model, where attention during inference is allocated to task-irrelevant regions. \textbf{(c)} An intuitive illustration: by progressively zooming in on task-critical regions, hallucinations (in red) diminish and the answers become correct. \textbf{(d)} Radar-chart comparisons among different reasoning models.}
    \label{fig:moti}
\end{figure*}


However, stronger reasoning can also exacerbate hallucination \cite{lu2025auditing,zhou2024mitigating,bai2024hallucination,wei2025memguard}, especially when long language-dominant deliberations encourage models to collect excessive or irrelevant clues \cite{wang2025vlm-r3,dong2025mirage,yi2025corgi,wang2025multimodal}, causing attention to drift away from visually decisive regions. We refer to this failure mode as \emph{reasoning drift} (Fig.~\ref{fig:moti}~(b)). As illustrated in Fig.~\ref{fig:moti}~(c), constraining the model toward task-critical regions can recover the correct answer, suggesting that the bottleneck lies not in model capacity but in ungrounded clue collection.

In practice, localizing task-critical visual clues at inference time faces two coupled difficulties:
(i) \textbf{long-horizon output control} and
(ii) \textbf{generalization beyond object-centric tasks}.
These difficulties stem from two implicit assumptions in existing mitigation strategies about where \textbf{clues} come from.
First, many methods are \emph{token-role-agnostic}: they apply global inference-time biasing, such as contrastive decoding, logit steering, or attention reallocation, without identifying which generated tokens require visual evidence~\cite{leng2024mitigating,im2025self,li2025hidden,tu2026attention,chen2025spatial}.
This can benefit short answers but may accumulate bias over long reasoning traces and miss clues that emerge only at intermediate steps~\cite{zheng2025lvlms}.
Second, other methods are \emph{task-agnostic}: they seek visual priors from richer captions or descriptions, assuming that the necessary clues can be obtained from the image alone~\cite{ghosh2024visual,yi2025corgi}.
Yet for complex reasoning tasks such as math~\cite{wang2024mathvision}, the question and intermediate reasoning often define which visual regions are actually decisive.
Thus, prior inference-time methods can struggle to isolate the \emph{decisive} clues required for long-chain multimodal reasoning.

To address these mismatches, we introduce \metric{}, a metric for measuring visual clue retrieval, and propose \method{}, a training-free inference plugin that suppresses hallucinations by tracing the flow of task-critical clues.
Rather than imposing a global visual bias or relying on image-only priors, \method{} explicitly follows a \textbf{query} $\rightarrow$ \textbf{output} $\rightarrow$ \textbf{vision} clue path.
It first uses the query to identify task constraints, then leverages generated tokens to determine which parts of the reasoning process need visual support, and finally traces these tokens back to the minimal supporting visual regions.
The traced regions are used to re-anchor subsequent reasoning to task-critical evidence, thereby mitigating hallucination without parameter updates or task-specific training.

Our contributions are as follows:
\begin{itemize}[noitemsep, topsep=0pt, leftmargin=*]
\item \textbf{(\metric{}) Layer-wise Average Clue Recall.} We introduce \metric{} to quantify layer-wise visual clue retrieval and reveal a consistent middle-to-late layer peak across architectures and scales.
\item \textbf{(\method{}) Hallucination suppression in reasoning architectures.} We propose \method{}, which traces native attention along \textbf{query} $\rightarrow$ \textbf{output} $\rightarrow$ \textbf{vision} to localize task-relevant evidence. It improves HallusionBench by \(\mathbf{1.25\times}\) and VMCBench by \(\mathbf{1.17\times}\).

\item \textbf{Transferability to non-reasoning MLLMs.} \method{} also transfers to non-reasoning MLLMs, yielding an average \textbf{+8.7} accuracy improvement on non-reasoning settings.
\end{itemize}

%% file: sections/3_related_work.tex
\section{Related Work}
\textbf{Multimodal Large Reasoning Models (MLRMs).} 
Recent MLLMs extend chain-of-thought reasoning to multimodal settings through supervised fine-tuning, reinforcement learning, search, reflection, self-refinement, and long-chain data scaling~\cite{zhao2024marco,zhang2024llamaberry,xu2025redstar,xu2024llava,thawakar2025llamav_o1,yao2024mulberry}. 
To improve reasoning reliability, preference-based and rule-based RL methods optimize models with human feedback, closed-loop refinement, reasoning trace comparisons, or GRPO-style rewards~\cite{yu2024rlhf,zhang2024improve,dong2024insight,MLRM-r1,liu2025segzero,xiao2025fast,huang2025vision,wang2025visualprm,meng2025mmeureka}. 
Existing MLRMs mainly follow two paradigms: two-stage SFT+RL pipelines and direct large-scale RL training for emergent reasoning~\cite{r1onevision,tan2025reason,zhang2025r1,ming2025oceanr1,wang2025sota}, with recent work further exploring unified reasoning-perception frameworks and RL-enhanced generative reasoning~\cite{ma2025one,huang2025gotr1}. 
However, how to keep long-chain multimodal reasoning grounded in task-critical visual evidence remains underexplored.

\begin{figure*}[htbp]
    \centering
    \includegraphics[width=\textwidth]{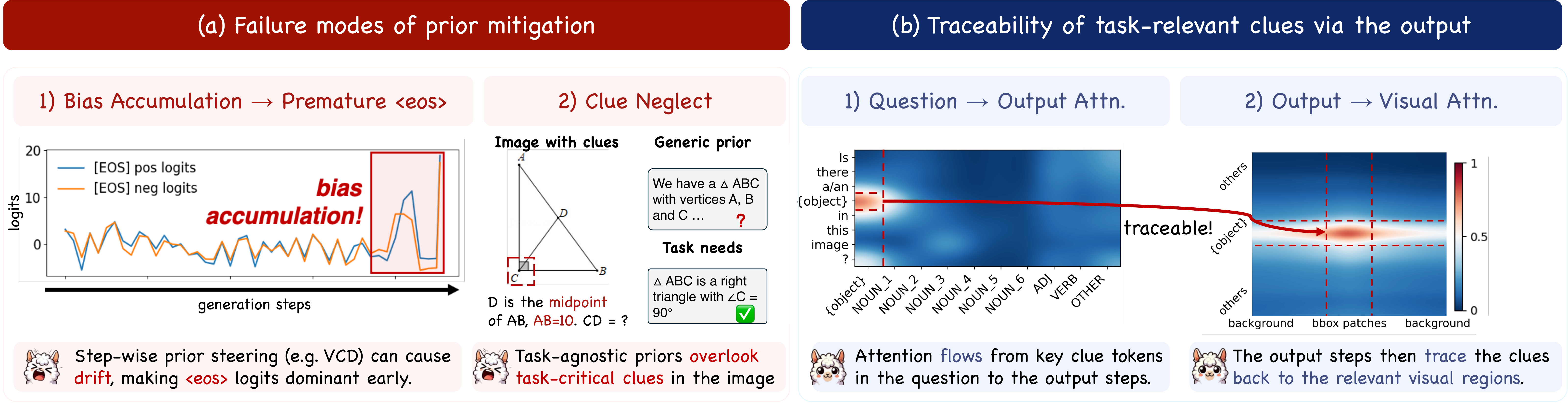}
    \caption{Our Motivations. From Failure Modes to Output-Guided Clue Tracing.}
    \vspace{-1em}
    \label{fig:obs1}
\end{figure*}

\noindent\textbf{Hallucination in MLRMs.} 
Long reasoning traces can amplify hallucination by encouraging models to rely on language priors over visual evidence~\cite{bai2024hallucination,liu2025amplified}. 
Existing inference-time mitigation methods either apply token-role-agnostic grounding, including contrastive decoding, logit steering, and attention control~\cite{leng2024mitigating,kim2024code,im2025self,anboosting,li2025hidden,kang2025see,tu2026attention,chen2025spatial,ye2025claim}, or introduce task-agnostic visual priors through generic descriptions or captions~\cite{ghosh2024visual,yi2025corgi}. 
The former may accumulate bias over long traces and miss decisive intermediate evidence~\cite{zheng2025lvlms}, while the latter often fails to capture question-conditional constraints required by complex reasoning tasks such as multimodal math~\cite{wang2024mathvision}. 
Although recent efforts improve hallucination mitigation through training-time alignment and datasets~\cite{dong2025mirage,wang2025vlm-r3,zhang2025focus,fan2025grit,zou2024memvr,wu2025mitigating,MLRM-aug-thinknot}, a step-aware, question-conditional inference-time mechanism for isolating \emph{decisive} visual clues is still lacking.

%% file: sections/4_method.tex
\section{Preliminaries}

\paragraph{Multimodal Large Reasoning Models (MLRMs).}
We consider decoder-style MLRMs that connect a pretrained visual encoder to a language model through a projector. 
The visual encoder and projector convert an image into visual tokens $\mathbf{X}_v$. 
Given system tokens $\mathbf{X}_s$ and question tokens $\mathbf{X}_q$, the multimodal prefix is
$
\mathbf{X}_c = [\mathbf{X}_s, \mathbf{X}_v, \mathbf{X}_q],
$
where $N_s$, $N_v$, and $N_q$ denote their lengths, and $N_c=N_s+N_v+N_q$. 
At decoding step $t$, the model predicts
$
y_t \sim p(y_t \mid \mathbf{X}_c,\mathbf{y}_{<t}),
$
where $\mathbf{y}_{<t}=\{y_i\}_{i=1}^{t-1}$. 
Reasoning-oriented MLLMs typically generate explicit reasoning traces, often enclosed by \texttt{<think>} and \texttt{</think>} tags, before producing the final answer~\cite{liu2023visual,zhang2023mmcot}.

\paragraph{Attention in MLRMs.}
For a Transformer decoder with $L$ layers, we denote by
$
\mathbf{A}_{t,l} \in [0,1]^{N_c+t-1}
$
the attention distribution at decoding step $t$ and layer $l$ over all available context tokens, including the multimodal prefix $\mathbf{X}_c$ and previously generated tokens $\mathbf{y}_{<t}$. 
Thus, each element $A_{t,l,i}$ measures how much the current decoding step attends to the $i$-th context token, and $\sum_i A_{t,l,i}=1$. 
This attention tensor allows us to aggregate token-level attention across layers and decoding steps.

\section{Motivations}
\label{sec:moti}
We conduct three targeted study to understand the thinking process of MLRMs and distill three observations that directly motivate our design in \method{}. Each observation is supported by a controlled study and empirical evidence.

\begin{obsbox}
\textbf{\textit{Obs 1:}} Bias accumulation and clue neglect in prior inference-time methods
\end{obsbox}
As shown in Fig.~\ref{fig:obs1} (a), directly steering output logits can encourage an MLRM to rely more on visual signals, but in long-chain reasoning it may induce \emph{bias accumulation}, which can prematurely trigger the \texttt{<eos>} token. For example, in VCD \cite{leng2024mitigating}, constructing a noisy negative counterpart and applying step-wise contrast can cause decoding to drift after a certain step, eventually making the \texttt{<eos>} logits dominant. Meanwhile, task-agnostic prior methods (e.g., generic captioning) still risk overlooking task-critical clues that are present in the image.
\begin{obsbox}
\textbf{\textit{Obs 2:}} Task-relevant clues are traceable across streams via the output
\end{obsbox}

To examine whether task-relevant clues can be traced across streams, we use category-bbox annotations to identify key visual regions and visualize the attention flow mediated by output tokens. 
Specifically, we compute attention heatmaps over 2,000 samples by tracing the chain from \textbf{question} to \textbf{output} to \textbf{vision}; implementation details are provided in Appendix~\ref{sec:app:moti}. 
As shown in Fig.~\ref{fig:obs1}~(b), the heatmaps reveal that decisive visual regions can be localized through this cross-stream chain, supporting the traceability of key clues via the output.



\begin{obsbox}
\textbf{\textit{Obs3:}} Key question tokens show larger \textbf{attn. variance} along the output dimension.
\end{obsbox}

Extending \textbf{\textit{Obs2}}, key question tokens exhibit localized attention peaks along the output axis; consequently their attention traces show larger variance than non-key tokens, providing a efficient, training-free signal for key-token detection.



These observations provide a pathway to suppress hallucinations in MLRMs: using the output as a conduit from the question to the visual tokens.

\section{Method}
\begin{figure*}[t]
    \centering
    \includegraphics[width=.95\textwidth]{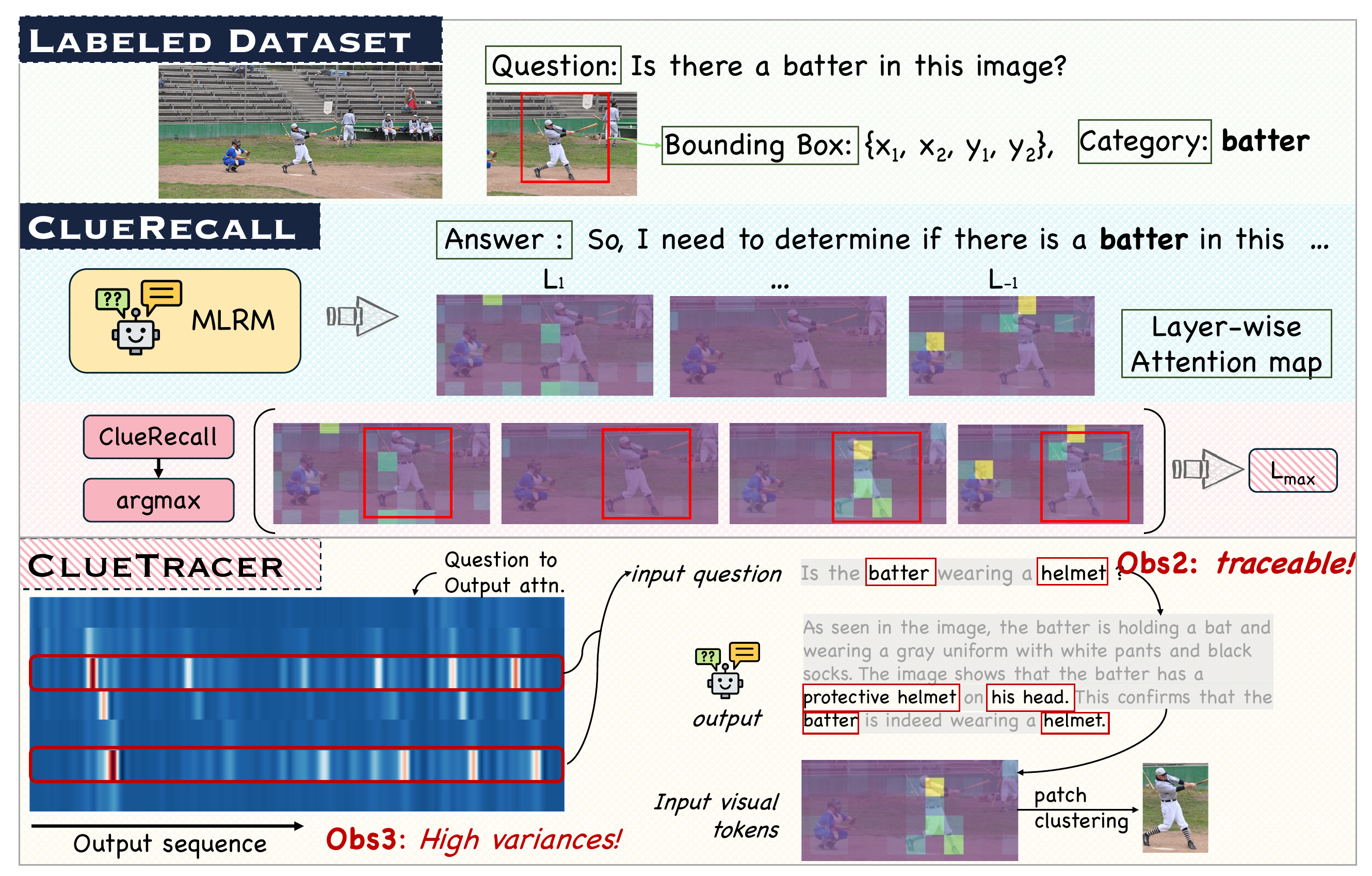}
    \caption{{Overview of our method. \textbf{Top}: \metric{} uses bbox/category annotations from COCO or lightweight detectors, enabling extension to arbitrary datasets. \textbf{Middle}: layer-wise \metric{} identifies $L_{\max}$, the layer with strongest visual clue retrieval. \textbf{Bottom}: \method{} traces the question $\rightarrow$ output $\rightarrow$ vision pathway to localize task-relevant regions. Best viewed in color.}}
    \label{fig:models}
    \vspace{-0.7em}
\end{figure*}

In this paper, we present \method{}, an inference-time approach that leverages an MLRM’s long reasoning chain to isolate visual clues critical for answering the question. We first introduce \metric{}, a layer-wise visual-grounding metric that identifies the decoder layer where clue retrieval is most reliable (Sec.~\ref{sec:a1}). Building on this, \method{} traces key clues across the model’s three streams, progressively localizing pivotal patch tokens via question $\rightarrow$ output $\rightarrow$ vision computations (Sec.~\ref{sec:key-patch-token}). Finally, we describe how the traced patches are organized into reusable clue regions for downstream reasoning (Sec.~\ref{sec:evidence-regions}). As shown in Fig.~\ref{fig:models}, \method{} uses stream-aware clue tracing to pinpoint the minimal visual support required for problem solving, thereby re-grounding generation and mitigating hallucinations.

\subsection{\metric{}: Assessing Visual Clue Retrieval}
\label{sec:a1}

{
Perception is a prerequisite for reliable reasoning. The goal of \metric{} is to assess, for each layer of an MLRM, how well its visual attention captures question-relevant visual clues. A single image typically contains multiple dense visual clusters, but only a subset is truly informative for answering a given question (\eg, when asked \texttt{``Is the batter wearing a helmet?''}, the decisive evidence should lie on the batter rather than the catcher). As a precursor to our method, \metric{} runs an automatic pipeline that measures how much attention each layer allocates to task-critical regions in three steps:
}

\paragraph{Query construction.}
{
For each image $I$, we obtain its set of bounding boxes and category labels $\texttt{cat}$ from either annotations (\eg, COCO \cite{lin2014microsoft}) or a segmentation model (\eg, Mask R-CNN \cite{he2017mask}). For each category, we instantiate a query using the fixed template $q_{\texttt{cat}} = \texttt{``Is there a/an \{\texttt{cat}\} in this image?''}$, and tokenize it into question tokens $\mathbf{X}_{q_\texttt{cat}}$.
}
\paragraph{Query-aligned regions.}
{
For each $(I,q_\texttt{cat})$ pair, we treat the corresponding bounding box of category $\texttt{cat}$ as the question-relevant clues. We encode $I$ into visual tokens $\mathbf{X}_v$ with the visual encoder and projector, and map the bounding box to its covered visual-token indices, denoted by $\mathbf{X}_{\texttt{bbox}} \subseteq \mathbf{X}_v$.
}
\paragraph{\metric{}.}
{
Collecting all such perception-labeled instances yields a set
$\mathcal{X}_{\texttt{perc}} = \{ (\mathbf{X}_{q_\texttt{cat}}, \mathbf{X}_v, \mathbf{X}_{\texttt{bbox}}, \texttt{cat}) \}$.
For each item in $\mathcal{X}_{\texttt{perc}}$, we prepend the system prompt $\mathbf{X}_s$ to form the complete input $\mathbf{X}_c = [\mathbf{X}_s, \mathbf{X}_v, \mathbf{X}_{q_\texttt{cat}}]$, feed it into the reasoning model $\mathcal{RM}$, and obtain output tokens $\{y_1,\dots,y_T\}$ and the attention tensor $\mathbf{A} \in [0,1]^{T \times L \times N_c}$. Let $\mathcal{V}$ denote the visual-token indices within $\mathbf{X}_c$, and let $\mathbf{A}_{t,l,\mathcal{V}} \in [0,1]^{N_v}$ be the restriction of $\mathbf{A}_{t,l}$ to these visual indices. We evaluate each layer $l$ by how well its visual attention, at the exact output steps where the category token is mentioned, retrieves the ground-truth region $\mathbf{X}_{\texttt{bbox}}$. For each layer $l \in \{1,\dots,L\}$, we define
}

\begin{equation}
\resizebox{\linewidth}{!}{$
\begin{aligned}
\mathrm{\textsc{ClueRecall}}(l)
&=
\frac{1}{|\mathcal{X}_{\texttt{perc}}|}
\sum_{\mathbf{X}_c \in \mathcal{X}_{\texttt{perc}}}
\\
&\quad
\frac{\left|
\mathrm{TopK}_{|\mathbf{X}_{\texttt{bbox}}|}
\left(
\sum_{t:\,y_t = \texttt{cat}} \mathbf{A}_{t,l,\mathcal{V}}
\right)
\cap \mathbf{X}_{\texttt{bbox}}
\right|}
{|\mathbf{X}_{\texttt{bbox}}|}.
\end{aligned}
$}
\label{eq:attnrecall}
\end{equation}

which directly measures a layer’s ability to retrieve object-aligned patches when the object is explicitly mentioned in the model’s output. Accordingly, we set
$L_{\max} \triangleq 
\operatorname*{arg\,max}\limits_{l \in \{1,\dots,L\}}
\mathrm{\textsc{ClueRecall}}(l)$
and use $L_{\max}$ as the reference layer for all subsequent clue-tracing computations.

\subsection{\method{}: Task-Critical Token Identification}
\label{sec:key-patch-token}


{
Building on \metric{}, we have identified the layer with the strongest \emph{clue retrieval}, denoted by $L_{\max}$. In \method{}, we move to a more general setting: we ask \emph{how to recover task-critical visual regions when neither bbox annotations nor fixed question templates are available}? To this end, we leverage the observations in Sec.~\ref{sec:moti} and use the pathway from question $\rightarrow$ output $\rightarrow$ visual tokens to progressively mine task-relevant clues.}

\paragraph{Key query tokens via output-axis variability.}
{
In \textbf{\textit{Obs~3}}, we find that when the model’s output is tightly aligned with the question, the decoder repeatedly allocates attention to a small set of key query tokens. As a result, the attention trajectory of these key tokens along the output axis becomes highly non-uniform, exhibiting larger variance. We therefore collect query-side attention at layer $L_{\max}$ across the entire output axis and retain query tokens whose standardized variability is high.}
{
Let $\mathcal{Q} \subseteq \{1,\dots,N_c\}$ denote the index set of query tokens within $\mathbf{X}_c$. For each query index $n_q \in \mathcal{Q}$ and each decoding step $t \in \{1,\dots,T\}$, we write}
\begin{equation}
\resizebox{\linewidth}{!}{$
A_{t,n_q} = A_{t,L_{\max},n_q},\;
\mathbf{a}_{n_q} = (A_{1,n_q},\dots,A_{T,n_q}) \in \mathbb{R}^{T}
$}
\label{eq:output2query}
\end{equation}
{
where $\mathbf{a}_{n_q}$ is the attention trajectory of the $n_q$-th query token $x_{n_q}$ over the output steps. We compute the variance $\mathrm{Var}(\mathbf{a}_{n_q})$ for each $n_q \in \mathcal{Q}$, and select key query tokens as
}
\begin{equation}
\mathcal{X}_q^{\star}
=
\bigl\{\, x_{n_q} \in \mathbf{X}_q \;:\; \text{zscore}(\mathrm{Var}(\mathbf{a}_{n_q})) \ge \tau_q \,\bigr\},
\label{eq:keyquery}
\end{equation}
{
where $\text{zscore}(\cdot)$ denotes the standard score obtained by subtracting the mean and dividing by the standard deviation \cite{zscore}. The indices in $\mathcal{X}_q^{\star}$ correspond to query tokens whose attention trajectories exhibit high output-axis variability and are thus treated as key query clues.
}

\begin{table*}[htbp]
\centering
\setlength{\tabcolsep}{5pt}
\renewcommand{\arraystretch}{1.15}
\caption{Comparison on reasoning-oriented hallucination benchmarks with and without \method{}. qAcc, fAcc, hAcc, and aAcc denote, respectively, the accuracy \emph{per question pair}, \emph{per figure}, \emph{on hard questions}, and the \emph{overall average}. Case studies are provided in Fig.~\ref{fig:appendix_case_01}-\ref{fig:appendix_case_04}.}
\resizebox{.95\linewidth}{!}{
\begin{tabular}{lccccccccccc}
\toprule
\multirow{2}{*}{\textbf{Model}} &
\multicolumn{4}{c}{\textbf{HallusionBench} (Accuracy \%)} &
\multicolumn{6}{c}{\textbf{VMCBench} (Accuracy \%)} &
\multirow{2}{*}{\shortstack[c]{\textbf{Average}\\ \textbf{$\Delta$Acc}}} \\
\cmidrule(lr){2-5}\cmidrule(lr){6-11}
 & qAcc & fAcc & hAcc & aAcc & Gen. & Reason. & OCR & Math & Doc\&Chart & Overall & \\ 
\midrule
\texttt{R1-OneVision} & 8.57 & 12.43 & 31.86 & 35.43 & 56.64 & 45.14 & 65.39 & 32.44 & 52.35 & 52.99 & \multirow{2}{*}{\shortstack[c]{\\ \textbf{+22.20\%}}} \\
\cellcolor{lightblue}  \textbf{+\method{}} & \cellcolor{lightblue} \textbf{18.18} & \cellcolor{lightblue} \textbf{30.79} & \cellcolor{lightblue} \textbf{41.16} & \cellcolor{lightblue} \textbf{49.16} & \cellcolor{lightblue} \textbf{76.11} & \cellcolor{lightblue} \textbf{61.81} & \cellcolor{lightblue} \textbf{89.97} & \cellcolor{lightblue} \textbf{51.21} & \cellcolor{lightblue} \textbf{78.88} & \cellcolor{lightblue} \textbf{73.91} &  \\
\midrule
\texttt{Ocean-R1} & 19.34 & 27.76 & 40.00 & 48.89 & 77.74 & 61.69 & 88.91 & 46.25 & 76.58 & 73.75 & \multirow{2}{*}{\shortstack[c]{\\ \textbf{+10.60\%}}} \\
\cellcolor{lightblue}  \textbf{+\method{}} & \cellcolor{lightblue} \textbf{30.11} & \cellcolor{lightblue} \textbf{43.35} & \cellcolor{lightblue} \textbf{54.88} & \cellcolor{lightblue} \textbf{63.51} & \cellcolor{lightblue} \textbf{80.31} & \cellcolor{lightblue} \textbf{62.72} & \cellcolor{lightblue} \textbf{96.24} & \cellcolor{lightblue} \textbf{52.00} & \cellcolor{lightblue} \textbf{81.07} & \cellcolor{lightblue} \textbf{77.02} &  \\
\midrule
\texttt{MM-Eureka} & 23.07 & 31.50 & 49.30 & 58.46 & 76.31 & 58.79 & 90.46 & 42.44 & 76.21 & 72.44 & \multirow{2}{*}{\shortstack[c]{\\ \textbf{+7.26\%}}} \\
\cellcolor{lightblue}  \textbf{+\method{}} & \cellcolor{lightblue} \textbf{33.41} & \cellcolor{lightblue} \textbf{47.11} & \cellcolor{lightblue} \textbf{56.98} & \cellcolor{lightblue} \textbf{65.10} & \cellcolor{lightblue} \textbf{83.56} & \cellcolor{lightblue} \textbf{67.26} & \cellcolor{lightblue} \textbf{94.05} & \cellcolor{lightblue} \textbf{58.38} & \cellcolor{lightblue} \textbf{85.93} & \cellcolor{lightblue} \textbf{80.31} &  \\
\midrule
\texttt{ORSTA-R1} & 21.76 & 28.61 & 45.81 & 55.45 & 71.55 & 55.62 & 85.06 & 42.82 & 72.60 & 68.39 & \multirow{2}{*}{\shortstack[c]{\\ \textbf{+7.30\%}}} \\
\cellcolor{lightblue} \textbf{+\method{}} & \cellcolor{lightblue}\textbf{27.91} & \cellcolor{lightblue}\textbf{36.71} & \cellcolor{lightblue}\textbf{53.49} & \cellcolor{lightblue}\textbf{60.05} & \cellcolor{lightblue}\textbf{81.52} & \cellcolor{lightblue}\textbf{64.21} & \cellcolor{lightblue}\textbf{94.41} & \cellcolor{lightblue}\textbf{51.11} & \cellcolor{lightblue}\textbf{84.63} & \cellcolor{lightblue}\textbf{78.39} &  \\
\bottomrule
\end{tabular}}
\label{tab:pope-vmc-a4}
\end{table*}

\paragraph{Tracing visual clues via Query-Output-Vision propagation.}
{
Having identified the key query clues $\mathcal{X}_q^{\star}$, we seek the \emph{task-supporting} visual evidence: a subset of visual tokens $\mathcal{X}_v^{\star} \subset \mathbf{X}_v$ that best \emph{grounds} these query constraints and thus suffices to solve the task. We formalize this as selecting $\mathcal{X}_v^{\star}$ to maximize the query-conditioned support $P(\mathcal{X}_v^{\star} \mid \mathcal{X}_q^{\star})$. However, under autoregressive (causally-masked) decoding, this dependency is not directly observable and is effectively mediated by the intermediate outputs, motivating a posterior signal. To bridge this, \method{} leverages the model's intermediate reasoning chain $\mathbf{y}_{1:T}$ as latent variables. We formulate the relevance of a visual token $x_v$ to the query constraints as a probabilistic trace, marginalizing over the output steps:
}
\begin{equation}
\resizebox{.88\columnwidth}{!}{$
P(x_v \mid \mathcal{X}_q^{\star}) 
= \sum_{t=1}^{T} 
\underbrace{P(x_v \mid y_t)}_{\text{Visual Grounding}} 
\cdot 
\underbrace{P(y_t \mid \mathcal{X}_q^{\star})}_{\text{Query Alignment}}.
$}
\label{eq:prob_trace}
\end{equation}
{
In this formulation, $P(y_t \mid \mathcal{X}_q^{\star})$ represents how strongly step $t$ aligns with the query constraints, and $P(x_v \mid y_t)$ represents the visual support required by that step.
In practice, we instantiate this probabilistic flow using the attention map at layer $L_{\max}$, which \metric{} identifies as the most reliable layer for \emph{clue retrieval}. We define the \emph{Trace Score} $\mathcal{T}_{\mathcal{RM}}(x_v)$ under the reasoning model $\mathcal{RM}$ by instantiating the probabilities with normalized attention weights:
}

\begin{equation}
\resizebox{\linewidth}{!}{$
\begin{aligned}
P(x_v \mid \mathcal{X}_q^{\star})
&\propto
\mathcal{T}_{\mathcal{RM}}(x_v) \\
&=
\sum_{t=1}^{T}
\underbrace{
\Bigl(
\sum_{x_{n_q} \in \mathcal{X}_q^{\star}}
A_{t,L_{\max},n_q}
\Bigr)
}_{\text{Query Alignment}}
\cdot
\underbrace{
A_{t,L_{\max},v}
}_{\text{Vis. Grounding}}.
\end{aligned}
$}
\label{eq:trace_impl}
\end{equation}

{
Eq.~\eqref{eq:trace_impl} effectively creates a continuous flow of importance from the query, through the reasoning bottleneck, to the image. High scores are assigned to visual tokens that are repeatedly attended to by reasoning steps that are themselves highly sensitive to the question. Finally, to suppress background noise, we apply a statistical filtering function $\psi(\cdot)$ (implemented via $z$-score normalization \cite{zscore}) to select the final task-critical set:
}
\begin{equation}
\mathcal{X}_v^{\star} = \bigl\{\, x_v \in \mathbf{X}_v \mid \psi(\mathcal{T}(x_v)) \ge \tau_v \,\bigr\}.
\label{eq:final_select}
\end{equation}
{
This traced set $\mathcal{X}_v^{\star}$ represents the minimal visual clues sufficient to support the query constraints, derived not from heuristics but from the model's dynamics.
}

\subsection{Evidence-Region Construction and Inference-Time Use}
\label{sec:evidence-regions}

\textbf{Organization.}
{
Starting from the selected visual tokens \(\mathcal{X}_v^{\star}\), we map each token \({x}_v^{\star} \in \mathcal{X}_v^{\star} \) to its image-plane center \(\phi({x}_v^{\star})\in\mathbb{N}^2\), cluster the point set \(\{\phi({x}_v^{\star})\}_{{x}_v^{\star}\in\mathcal{X}_v^{\star}}\) with DBSCAN \cite{ester1996dbscan} to obtain clusters \(\{C_{r}\}_{r=1}^{R}\), and enclose each cluster by an axis-aligned rectangle \(R_{r}=\mathrm{BBox}(C_{r})\). The resulting region set \(\mathcal{R}=\{R_{1},\ldots,R_{R}\}\) defines crops of the original image. This construction preserves the model’s native rectangular interface while consolidating fragmented evidence into a small number of spatially coherent \emph{visual clues}.
}

\textbf{Input modes.}
We use these crops in two complementary ways. \emph{Offline:} precompute and cache \(\mathcal{R}\) for repeated evaluation or families of similar images. \emph{Online:} a two-stage inference pipeline—\textbf{Stage~1} derives \(\mathcal{R}\) from the model; \textbf{Stage~2} re-invokes the model on \((I;\mathcal{R})\) to refine reasoning with focused evidence. 

%% file: sections/5_experiment.tex
\begin{table*}[htbp]
\centering
\caption{Per-layer \metric{} (left) and accuracy on \textsc{POPE}/\textsc{MMVP} with and without \method{} (right).}
\label{tab:perc_a4}
\setlength\tabcolsep{4pt}
\resizebox{\linewidth}{!}{
\begin{tabular}{lcccccc|cc|cc}
\hline
& \multicolumn{6}{c|}{\textbf{\metric{} per Layer (\%)}} & \multicolumn{2}{c|}{\textbf{POPE (\%)}} & \multicolumn{2}{c}{\textbf{MMVP (\%)}} \\
\textbf{Model} & \textbf{0} & \textbf{6} & \textbf{12} & \textbf{18} & \textbf{24} & \textbf{27} & \textbf{w/o \textsc{CT}} & \textbf{w/ \textsc{CT}} & \textbf{w/o \textsc{CT}} & \textbf{w/ \textsc{CT}} \\
\hline
\texttt{R1-OneVision}  & 30.78\% & 33.68\% & 43.11\% & \textbf{50.60\%} & 47.28\% & 44.11\% & 70.22\% & \cellcolor{lightblue}{\textbf{76.97\%}} & 23.33\% & \cellcolor{lightblue}{\textbf{46.00\%}} \\
\texttt{Ocean-R1}         & 31.05\% & 32.64\% & 48.56\% & \textbf{53.79\%} & 50.92\% & 49.23\% & \textbf{86.77\%} & \cellcolor{lightblue}{85.77\%} & 47.33\% & \cellcolor{lightblue}{\textbf{50.00\%}} \\
\texttt{MM-Eureka}     & 29.30\% & 30.57\% & 45.45\% & 51.06\% & \textbf{51.38\%} & 48.11\% & 75.00\% & \cellcolor{lightblue}{\textbf{81.38\%}} & 44.67\% & \cellcolor{lightblue}{\textbf{50.00\%}} \\
\texttt{ORSTA-R1}         & 31.14\% & 32.38\% & 47.55\% & 51.66\% & \textbf{55.32\%} & 50.56\% & 71.36\% & \cellcolor{lightblue}{\textbf{82.98\%}} & 38.00\% & \cellcolor{lightblue}{\textbf{53.33\%}} \\
\hline
\end{tabular}
}
\end{table*}

\begin{figure*}[tb]
    \centering
    \includegraphics[width=.90\textwidth]{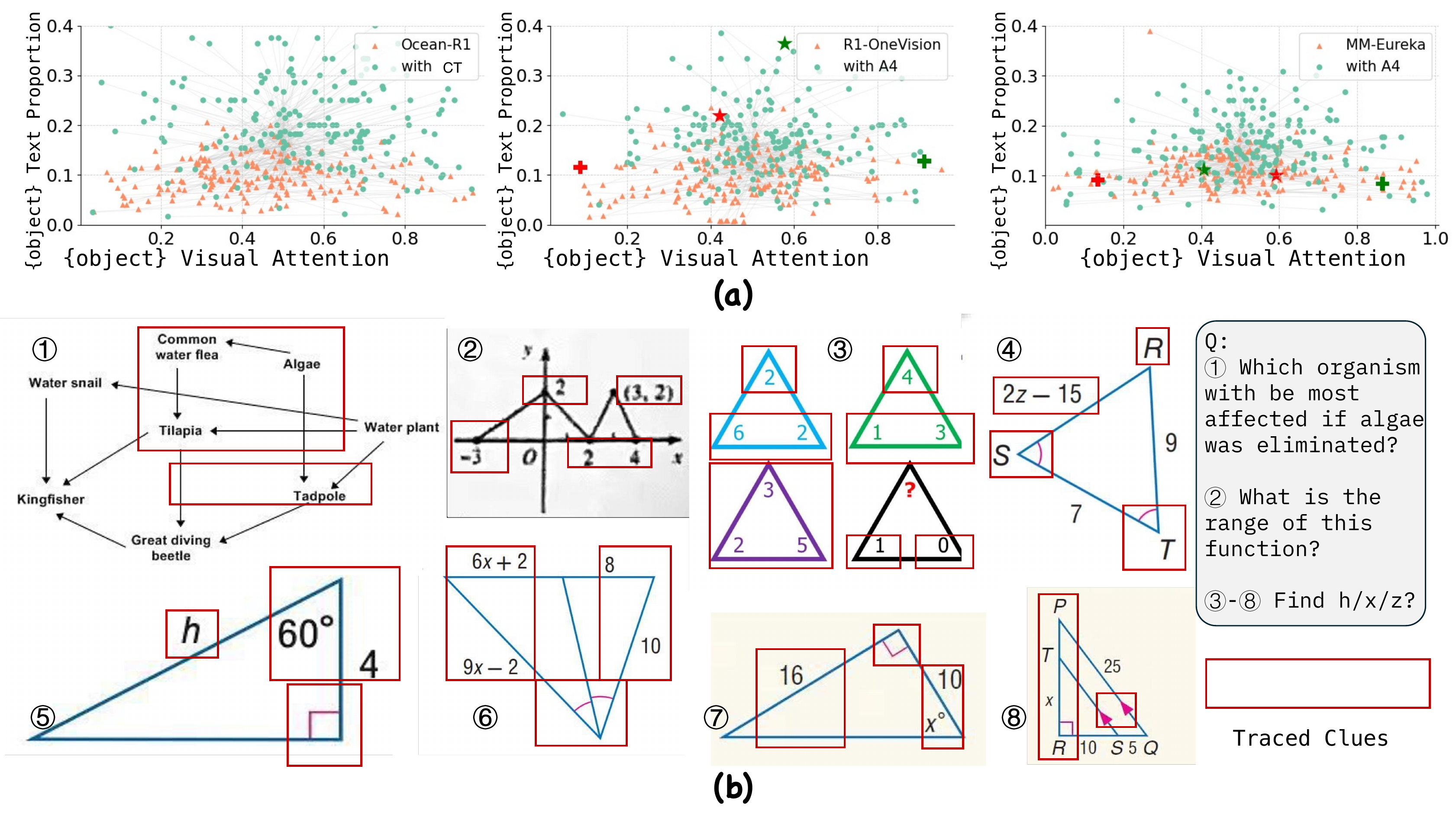}
    \caption{
Visualization of reduced reasoning drift and traced visual clues. 
\textbf{(a)} Comparative scatter plot of attention allocation versus output mentions for task-relevant objects. 
\textbf{(b)} Case studies showing that \method{} traces implicit visual clues required by the question, with red boxes highlighting the discovered conditions in the image.
}
    \vspace{-1.1em}
    \label{fig:rq2}
\end{figure*}

\section{Experiments}
In this section, we conduct experiments to address the following research questions:

\begin{itemize}[noitemsep, topsep=0pt, leftmargin=*]
    \item \textbf{RQ1.} How well does \method{} reduce hallucinations and improve accuracy across diverse reasoning architectures? Which layer achieves the highest \metric{}?
    \item \textbf{RQ2.} During generation, can \method{} mitigate reasoning drift exhibited in Fig.~\ref{fig:moti}, maintaining alignment between the question focus and produced content?
    \item \textbf{RQ3.} How do results change when key settings varied, and specifically, does \method{}’s output-mediated patch selection outperform baselines designed for non-reasoning MLLMs?
\end{itemize}

\subsection{Experimental Setup}
\label{sec:exp_setup}

We begin with a concise overview of models, datasets, and evaluation metrics; full implementation details are deferred to the Appendix~\ref{sec:app_exp_detailed}.

\noindent\textbf{Models.} \method{} is compatible with existing hallucination-mitigation schemes for MLRMs. Accordingly, we apply it to several reasoning models—\texttt{R1-OneVision} \cite{r1onevision}, \texttt{Ocean-R1}~\cite{ming2025oceanr1}, \texttt{Orsta-R1}~\cite{ma2025one}, and \texttt{MM-Eureka}~\cite{meng2025mmeureka}—to assess generality. To test transferability, we also port \method{} to \emph{non-reasoning} MLLMs, including \texttt{LLaVA-1.6}~\cite{liu2024llavanext} and \texttt{Qwen2.5-VL}~\cite{Qwen2.5-VL}, as discussed in Appendix~\ref{sec:non-reasoning}.

\noindent\textbf{Datasets \& Evaluation Metrics.} 
We evaluate on four widely used hallucination benchmarks, grouped as follows: \emph{reasoning-oriented}: \textbf{1)} \textbf{VMCBench}~\cite{Zhang_2025_CVPR}, a unified multiple-choice questions drawn from $20$ VQA datasets, e.g., \emph{MathVision}~\cite{wang2024mathvision}, \emph{ScienceQA}~\cite{lu2022scienceqa}, and \textbf{2)} \textbf{HallusionBench}~~\cite{Guan_2024_CVPR} benchmarks image–context reasoning where language hallucinations and visual illusions are entangled; and \emph{perception-oriented}: \textbf{3)} \textbf{POPE}~\cite{li2023pope} and \textbf{4)} \textbf{MMVP}~\cite{tong2024mmvp}. We follow each dataset’s official evaluation protocol.

\subsection{Performance with \method{} on Reasoning Models (RQ1)}
 To assess hallucination in \emph{reasoning} models, we evaluate each model \emph{before} and \emph{after} applying \method{} on hallucination benchmarks; the results are reported in Table~\ref{tab:pope-vmc-a4} and Table \ref{tab:perc_a4}, Additional results on \textbf{other variants} and \textbf{larger models} are provided in Appendix~\ref{sec:app:detailed_res}. We can draw the following findings: 

\begin{figure*}[htbp]
\centering
\small
\setlength{\tabcolsep}{3.5pt}
\setlength{\aboverulesep}{0.35ex}
\setlength{\belowrulesep}{0.35ex}
\setlength{\cmidrulekern}{0.25em}

\newlength{\panelH}
\setlength{\panelH}{4.0cm} 

\begin{tabular}{@{} p{0.36\textwidth} p{0.25\textwidth} p{0.315\textwidth} @{}}
\begin{minipage}[t][\panelH][t]{\linewidth}
  \centering
  \textbf{(a) Attention Layer}\par
  \vspace{0.5\baselineskip}
  \includegraphics[width=\linewidth, height=4cm, keepaspectratio]{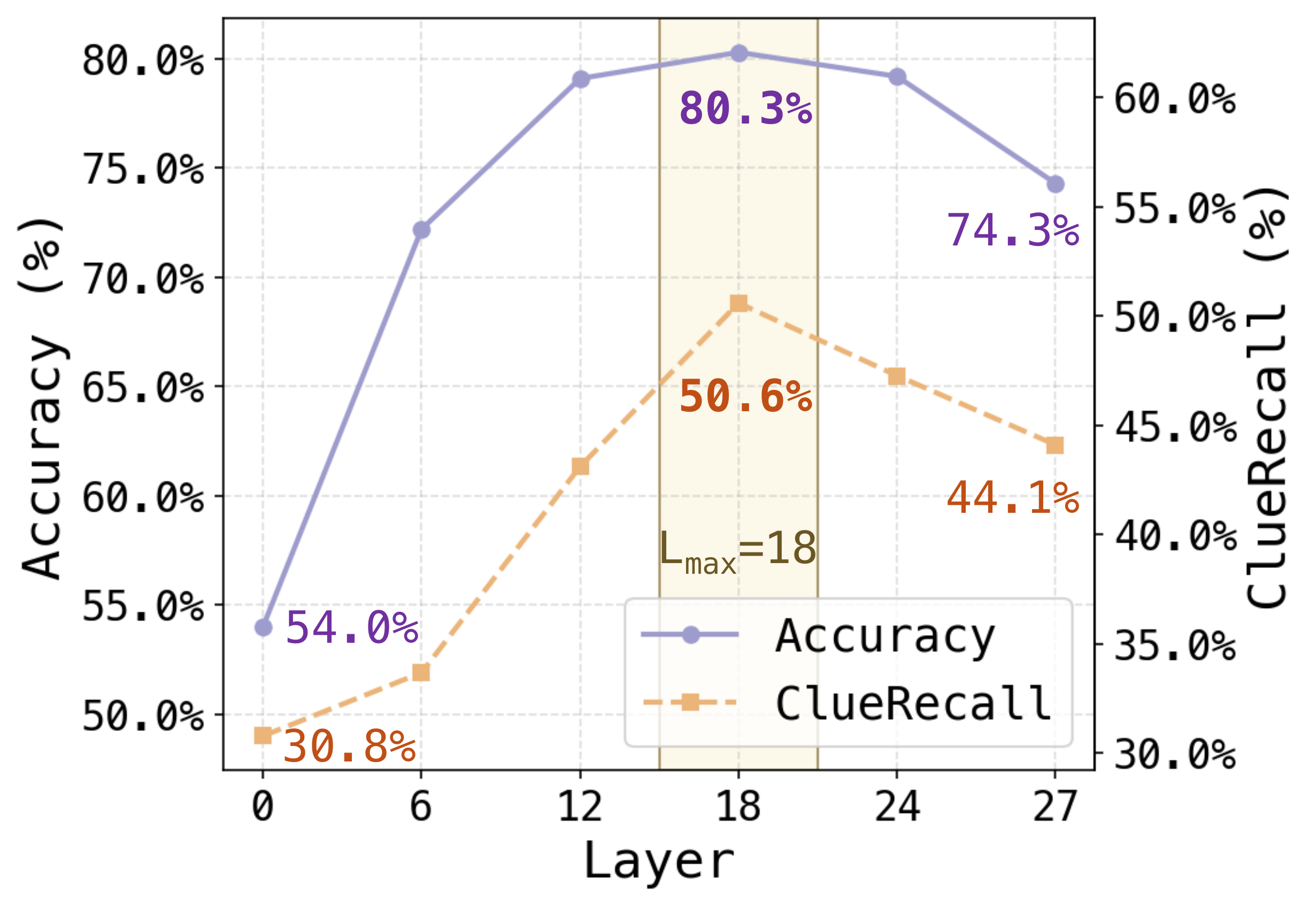}
\end{minipage}
&
\begin{minipage}[t][\panelH][t]{\linewidth}
  \centering
  \textbf{(b) Components}\par
  \vspace{0.5\baselineskip}
  \renewcommand{\arraystretch}{1.025}
  \resizebox{\linewidth}{!}{%
  \begin{tabular}{lc}
    \toprule
    \textit{Attentions} & Acc \\
    \midrule
    output-visual    & 72.5\% \\
    question-visual  & 48.8\% \\
    \textbf{Ours}    & \textbf{80.3}\% \\
    \midrule
    \multicolumn{2}{l}{\textit{Clustering Method}} \\
    \midrule
    \textbf{DBSCAN}  & \textbf{80.3\%} \\
    K-means          & 79.2\% \\
    \bottomrule
  \end{tabular}}
\end{minipage}
&
\begin{minipage}[t][\panelH][t]{\linewidth}
  \centering
  \textbf{(c) Accuracy vs. Time}\par
  \vspace{0.5\baselineskip}
  \renewcommand{\arraystretch}{1.05}
  \resizebox{\linewidth}{!}{%
  \begin{tabular}{lcc}
    \toprule
    Method & Time ($\downarrow$) & Acc. ($\uparrow$) \\
    \midrule
    Vanilla & 0.9h & 73.8\% \\
    VCD     & 1.3h & 68.1\% \\
    ICD    & 1.4h & 64.2\% \\
    AGLA   & 1.4h & 66.3\% \\
    VDGD   & 1.9h & 74.1\% \\
    Ours    & 1.8h & \textbf{80.3\%} \\
    \bottomrule
  \end{tabular}}
\end{minipage}
\\
\end{tabular}
\caption{Ablation Study and Comparative Evaluations. Default settings in the experiments are shown in \textbf{bold}. Additional ablations on hyperparameters (e.g., DBSCAN and z-score thresholds) are provided in the Appendix \ref{subsec:more-sensitivity}.}
\label{tab:ablation-study}
\vspace{-1em}
\end{figure*}

\noindent\textbf{Finding 1: Consistent gains across models and datasets.}
\method{} improves \emph{all} reasoning models across the four hallucination benchmark. Specifically, on the reasoning-oriented hallucination datasets—\textbf{HallusionBench} and \textbf{VMCBench}—\method{} achieves a \textbf{maximum} accuracy gain of \textbf{22.20\%} and an \textbf{average} accuracy gain of \textbf{11.84\%}.

\noindent\textbf{Finding 2: Models with different training architectures exhibit similar perceptual capacity.}
From Table~\ref{tab:perc_a4}, the \metric{} curves for all models peak around \textbf{50\%--75\%} of the network depth and follow a rise--then--fall pattern. 
Appendix~\ref{subsec:more-sensitivity} further shows that this phenomenon remains consistent across a broader range of architectures and model scales.
\subsection{Hallucination Analysis (RQ2)}
To further verify the hallucination–suppression ability of \method{}, we compute two key indicators on the \textsc{POPE} dataset: (1) \{\textit{object}\} \emph{visual attention}, i.e., the attention mass that falls inside the \{\textit{object}\} bounding box; and (2) \{\textit{object}\} \emph{text proportion}, i.e., the fraction of noun tokens in the model output that correspond to \{\textit{object}\}, We further provide MathVista \cite{lu2023mathvista} examples in Fig.~\ref{fig:rq2}~(b), where many decisive clues, such as parallelism and perpendicularity, must be grounded in the image. Our key findings are as follows:

\noindent\textbf{Finding 3: \method{} mitigates reasoning drift.}
As shown in Fig.~\ref{fig:rq2}~(a), after models are equipped with \method{}, the \{\textit{object}\} \emph{text proportion} in the generated text increases markedly across all models. 
This indicates that the reasoning process pays more attention to task-critical content, thereby reducing reasoning drift. 
Fig.~\ref{fig:rq2}~(b) further provides MathVista examples, showing that \method{} can localize implicit visual clues required by the question, such as \textit{parallelism, perpendicularity, and equal angles}, even when the question is as simple as ``find \(x\).''

\subsection{Ablation and Comparison (RQ3)}
\label{sec:ablat}
\makeatletter
\newcommand\figcaption{\def\@captype{figure}\caption}
\newcommand\tabcaption{\def\@captype{table}\caption}
\makeatother

As discussed in pervious sections, \method{} effectively identifies task-relevant visual evidence. In this section, we examine the \emph{module sensitivity} of \method{} and explain why analogous approaches developed for \emph{non-reasoning} MLLMs do not carry over to \emph{reasoning} models. We conduct ablation studies and comparative evaluations on \texttt{Ocean-R1} using VMCBench. For the comparative setting, we benchmark against prior contrastive-decoding and description-prior methods:

\textbf{VCD} \cite{leng2024mitigating} contrasts decoding distributions from the original image and a corrupted view; \textbf{ICD} \cite{wang2024mitigating} contrasts standard vs. disturbance-instruction decoding to subtract hallucinated concepts; \textbf{AGLA} \cite{an2025mitigating} assembles global generation with prompt-relevant local cues via an augmented (masked) view; \textbf{VDGD} \cite{ghosh2024visual} grounds decoding with a self-generated visual description prefix to better support reasoning. The results are summarized in Fig.~\ref{tab:ablation-study}. Our findings are:

\noindent\textbf{Finding 4: Performance trends align with expectations.}
Based on Figure~\ref{tab:ablation-study}~(a–b), the accuracy trend over attention layers mirrors the \metric{} pattern in Table~\ref{tab:perc_a4}. 
{For \textbf{clustering}, DBSCAN is preferable to K\mbox{-}means, as it is less sensitive to noise and can naturally discard isolated noisy tokens when the attention map contains spurious activations}; see Appendix~\ref{sec:app:dis} for a detailed discussion.

\noindent\textbf{Finding 5: \method{} outperforms prior inference-time baselines on reasoning settings.}
As shown in Fig.~\ref{tab:ablation-study}~(c), \method{} consistently improves over baselines originally developed for non-reasoning settings, which are less effective under the long-context dynamics of reasoning models. In particular, contrastive-decoding methods can suffer from \emph{bias accumulation} over long chains, causing decoding to drift and even reducing accuracy below the vanilla model, while description-prior approaches often fail to provide task-specific priors that are actually useful for solving the question.

%% file: sections/6_conclusion.tex
\section{Conclusion}
In this paper, we introduce \method{}, a training-free approach for suppressing hallucination in MLRMs. 
\method{} traces the question$\rightarrow$output$\rightarrow$vision attention pathway to localize task-relevant visual patches and guide inference toward decisive evidence. 

\section*{Limitations}

While \method{} efficiently mitigates hallucination in MLRMs, several limitations merit acknowledgment and point to concrete remedies. 
\textbf{(i) Two-stage inference.} Owing to current reasoning-model architectures, \method{} operates in two stages (attention mining then focused re-inference). A learned controller that adaptively decides \emph{when} to trigger \method{} and \emph{how much} context to generate (e.g., output length for Stage~1) could reduce overhead and retain quality in a single pass. 
\textbf{(ii) Rectangular crops.} Our evidence regions are axis-aligned; objects, however, are often non-rectangular, so rectangular cropping may include extraneous background.

\paragraph{Future work.}
\textbf{(i) Single-pass integration.} Post-train a lightweight policy to fuse \method{} into one forward pass, enabling adaptive, example-dependent activation and scope (triggering and output length) within the model’s native decoding; reinforcement learning or adaptive-compute objectives are natural fits. 
\textbf{(ii) Beyond rectangles.} Explore mask- or token-level selection: derive instance/part masks with promptable segmentation to form shape-accurate evidence, or re-inject only selected visual tokens (retaining their original positional encodings) while dropping or down-weighting others, thus avoiding commitment to a particular crop geometry.
\textbf{(iii) Compute-network-aware deployment.} 
Given its training-free and plug-and-play nature, \method{} provides a promising direction for resource-constrained or collaborative edge-cloud settings~\cite{cheng2023dynamic}. Specifically, Stage 1 could be replaced by a lightweight clue-tracing model deployed on edge nodes, which efficiently identifies task-relevant visual evidence and forwards the extracted clues to a cloud-based large model for reasoning. Such a design can enable flexible compute-network resource orchestration across heterogeneous infrastructures, effectively reducing inference costs while preserving or even enhancing reasoning accuracy in distributed deployment scenarios.

\section*{ACKNOWLEDGMENTS}
This work was supported in part by the National Key Research and Development Program of China under Grant 2024YFB2906705 and the National Natural Science Foundation of China under Grant 62676038.


%% file: sections/7_appendix.tex
\clearpage
\begin{table*}[htbp]
\centering
\caption{\textbf{{Notations used throughout \metric{} and \method{}.}}}
\label{tab:notation}
\resizebox{\linewidth}{!}{%
\begin{tabular}{ll}
\toprule
Notation & Description \\
\midrule
$\mathcal{RM},\, L$ & Multimodal reasoning model (MLRM) and its number of Transformer layers \\
$I,\, q$ & Input image and textual que \\
$\texttt{cat},\, \texttt{bbox}$ & Category of an entity in $I$ and its bounding box in the image plane \\
$\mathbf{X}_c,\, \mathbf{X}_s,\, \mathbf{X}_q,\, \mathbf{X}_v$ 
& Complete input, system prompt tokens, query tokens, and visual tokens \\
$N_c,\, N_s,\, N_q,\, N_v$ 
& Lengths of $\mathbf{X}_c$, $\mathbf{X}_s$, $\mathbf{X}_q$, and $\mathbf{X}_v$ \\
$\mathbf{X}_{q_{\texttt{cat}}},\, \mathbf{X}_{\texttt{bbox}}$ 
& Query tokens for category \texttt{cat}; visual tokens within the bounding box \texttt{bbox} \\
$\mathcal{X}_{\text{perc}}$ 
& Labeled dataset of $(\mathbf{X}_{q_\texttt{cat}}, \mathbf{X}_v, \mathbf{X}_{\texttt{bbox}}, \texttt{cat})$ for computing \metric{}  \\
$\mathbf{y}_{1:T}$ & Model output token sequence $\{y_1,\dots,y_T\}$ of length $T$ \\
$\mathbf{A}$ & Attention tensor over inputs, $\mathbf{A} = (A_{t,l,n}) \in [0,1]^{T \times L \times N_c}$ \\
$\mathcal{X}_q^{\star},\, \mathcal{X}_v^{\star}$ 
& Key query tokens and task-critical visual tokens traced by \method{} \\
$\mathcal{T}(x_v)$ & Trace score representing the probability flow $P(x_v|\mathcal{X}_q^{\star})$ via output outputs \\
$\psi(\cdot),\, \phi(\cdot)$ 
& Filtering function (implemented as $z$-score); mapping from a region to its bounding box \\
\bottomrule
\end{tabular}%
}
\end{table*}
\section{LLM Usage}
We used a large language model \textbf{[ChatGPT 5.4]} strictly for writing assistance limited to spelling, grammar, and minor stylistic polishing.
\section{Notations}
The notation used in this paper is summarized in Table~\ref{tab:notation}.

\section{Experiment Details}
\label{sec:app_exp_detailed}

\subsection{Baselines}

\paragraph{Models.}
Because \method{} is training-free and architecture-agnostic, we integrate it at inference into four recent multimodal reasoning models without altering their weights or decoding. (i) \textbf{R1-OneVision}~\cite{r1onevision}—a cross-modal formalization pipeline that converts visual content into structured textual representations before language reasoning, trained with SFT and RL. (ii) \textbf{Ocean-R1}~\cite{ming2025oceanr1}—an open RL-enhanced VLM whose public artifacts indicate a two-stage recipe (reasoning first, then visual perception) with released weights/data. (iii) \textbf{Orsta}~\cite{ma2025one}—a family of VLMs post-trained via the \emph{V-Triune} unified RL system to improve both perception and reasoning. (iv) \textbf{MM-Eureka}~\cite{meng2025mmeureka}—a rule-based (R1-style) reinforcement learning approach extended to multimodal reasoning with an open pipeline. To assess transferability, we further deploy the \method{}-derived visual cropping prior on two \emph{non-reasoning} MLLMs: (v) \textbf{Qwen2.5-VL}~\cite{Qwen2.5-VL}—a general-purpose vision–language model series (3B/7B/72B) with an efficient ViT and official instruction-tuned checkpoints; and (vi) \textbf{LLaVA-1.6-Mistral}~\cite{liu2024llavanext}—an instruction-tuned VLM pairing a vision encoder with Mistral-7B, documented with improved training data and dynamic high-resolution support.

\paragraph{Datasets.}
We evaluate hallucination on four benchmarks. \textbf{POPE}~\cite{li2023pope} targets object-level hallucination via yes/no “object presence” queries and contains three standard splits (popular/adversarial/random). \textbf{VMC-Bench}~\cite{Zhang_2025_CVPR} unifies 20 VQA datasets into a single multiple-choice suite (9{,}018 questions) spanning diverse domains. \textbf{MMVP}~\cite{tong2024mmvp} focuses on nine basic visual patterns and “CLIP-blind pairs,” probing failure modes that induce incorrect answers or overconfident, hallucinatory explanations. \textbf{HallusionBench}~\cite{Guan_2024_CVPR} is an image–context reasoning benchmark diagnosing entangled language hallucination and visual illusion, consisting of 346 images and 1,129 expert-crafted questions with control structures for analysis.
For \textbf{POPE} we report accuracy using the \emph{lmms-eval}~\cite{zhang2024lmmsevalrealitycheckevaluation} implementation. For \textbf{VMC-Bench}, \textbf{MMVP}, and \textbf{HallusionBench} we follow each benchmark’s official evaluation pipeline.

\paragraph{License.} All models evaluated in this work are publicly available open-weight models. Specifically, \texttt{R1-OneVision}, \texttt{MM-Eureka}, \texttt{LLaVA-1.6}, and \texttt{Qwen2.5-VL} are released under the Apache-2.0 license, while \texttt{Orsta-R1} is released under the MIT license. \texttt{Ocean-R1} is publicly released through its official Hugging Face and GitHub repositories, but we did not identify a separate license statement for its model weights. We use all models only for academic evaluation and do not redistribute model weights.

\subsection{Evaluation setup details}

\label{sec:expsetup}

\noindent\textbf{Prompt templates.} For \emph{yes/no} questions (POPE, MMVP, HallusionBench) and \emph{multiple choice} questions (VMCBench), we use the following templates:

\begin{obsbox}
You FIRST think about the reasoning process as an internal monologue and then provide the final answer.
The reasoning process MUST BE enclosed within <think></think> tags. The final answer MUST BE a single word enclosed in <answer></answer> tags. Let's think more.\\
\{question\}
\end{obsbox}

\begin{obsbox}
You FIRST think about the reasoning process as an internal monologue and then provide the final answer.
The reasoning process MUST BE enclosed within <think></think> tags. The final answer must be an option letter from the given choices, enclosed in <answer></answer> tags. Let's think more.\\
\{question\}
\end{obsbox}

\noindent\textbf{Evaluation.} During evaluation, we parse the model output with a regular expression to extract the content enclosed in \texttt{<answer>...</answer>} and compare it with the ground truth.

\noindent\textbf{Generate config.} We use the same decoding settings across all datasets.
\begin{table}[h]
\centering
\begin{tabular}{ll}
\toprule
\textbf{Parameter} & \textbf{Value} \\
\midrule
\texttt{min\_new\_tokens} & 50 \\
\texttt{max\_new\_tokens} & 1024 \\
\texttt{do\_sample} & True \\
\bottomrule
\end{tabular}
\end{table}
\section{More Experimental Results}
\label{sec:app:detailed_res}
\subsection{More results on POPE and VMC-Bench}
\paragraph{Evaluation Setup.}
We report detailed results across multiple benchmarks following each dataset’s official protocol. For \textsc{POPE}, we report accuracy and F1 on the \emph{Random}, \emph{Popular}, and \emph{Adversarial} splits \cite{li2023pope}. For \textsc{VMCBench}, we report per\mbox{-}subset and overall accuracy \cite{Zhang_2025_CVPR}. For \textsc{MMVP}, we follow the authors’ evaluation on CLIP\mbox{-}blind visual patterns \cite{tong2024mmvp}.

\paragraph{\textsc{VMCBench} Subsets.}
We adopt \textsc{VMCBench}’s four capability\mbox{-}oriented subsets and their official composition \cite{Zhang_2025_CVPR}:
\textbf{(1) General}: \emph{VQAv2}~\cite{goyal2017making}, \emph{GQA}~\cite{hudson2019gqa}, \emph{VizWiz}~\cite{gurari2018vizwiz}, \emph{OK\mbox{-}VQA}~\cite{marino2019okvqa}, \emph{A\mbox{-}OKVQA}~\cite{schwenk2022aokvqa}; 
\textbf{(2) Reasoning}: \emph{MMMU}~\cite{yue2023mmmu}, \emph{MathVista}~\cite{lu2023mathvista}, \emph{MathVision}~\cite{wang2024mathvision}, \emph{MMStar}~\cite{chen2024mmstar}; 
\textbf{(3) OCR}: \emph{TextVQA}~\cite{singh2019textvqa}, \emph{OCR\mbox{-}VQA}~\cite{mishra2019ocrvqa}; 
\textbf{(4) Doc \& Chart}: \emph{DocVQA}~\cite{mathew2021docvqa}, \emph{TableVQA\mbox{-}Bench}~\cite{kim2024tablevqabench}, \emph{InfographicVQA (InfoVQA)}~\cite{mathew2022infographicvqa}, \emph{AI2D}~\cite{kembhavi2016diagram}, \emph{ChartQA}~\cite{masry2022chartqa}.

\begin{table*}[t]
\centering
\setlength{\tabcolsep}{4pt}
\renewcommand{\arraystretch}{1.12}
\caption{Accuracy (\%) on selected \textsc{VMCBench}. Including: MMMU, MathVista, MMStar, AI2D, ScienceQA, SEEDBench, MM-Vet, MathVision, TableVQABench (TVQA), RealWorldQA (RWQA). Rows are paired as baseline vs.\ +\method{} for each model.}
\label{tab:vmc-split-a}
\resizebox{\linewidth}{!}{
\begin{tabular}{lccccccccccc}
\toprule
\textbf{Model} & \textbf{MMMU} & \textbf{MathVista} & \textbf{MMStar} & \textbf{AI2D} & \textbf{ScienceQA} & \textbf{SEEDBench} & \textbf{MM-Vet} & \textbf{MathVision} & \textbf{TVQA} & \textbf{RWQA} \\
\midrule
R1\mbox{-}OneVision & 36.30 & 36.63 & 40.38 & 46.70 & 61.54 & 54.81 & 48.12 & 28.99 & 48.31 & 46.79 \\
\cellcolor{lightblue}\textbf{+\method{} (Ours)} & \cellcolor{lightblue}\textbf{50.49} & \cellcolor{lightblue}\textbf{67.00} & \cellcolor{lightblue}\textbf{55.96} & \cellcolor{lightblue}\textbf{68.06} & \cellcolor{lightblue}\textbf{82.35} & \cellcolor{lightblue}\textbf{72.35} & \cellcolor{lightblue}\textbf{74.29} & \cellcolor{lightblue}\textbf{36.41} & \cellcolor{lightblue}\textbf{73.79} & \cellcolor{lightblue}\textbf{54.62} \\
\midrule
Ocean\mbox{-}R1 & 56.28 & 60.89 & 58.81 & 74.94 & 80.09 & 73.58 & 73.85 & 32.35 & 69.02 & 60.09 \\
\cellcolor{lightblue}\textbf{+\method{} (Ours)} & \cellcolor{lightblue}\textbf{55.42} & \cellcolor{lightblue}\textbf{69.50} & \cellcolor{lightblue}\textbf{62.53} & \cellcolor{lightblue}\textbf{79.68} & \cellcolor{lightblue}\textbf{84.39} & \cellcolor{lightblue}\textbf{80.99} & \cellcolor{lightblue}\textbf{79.44} & \cellcolor{lightblue}\textbf{36.97} & \cellcolor{lightblue}\textbf{79.09} & \cellcolor{lightblue}\textbf{66.67} \\
\midrule
MM\mbox{-}Eureka & 51.20 & 58.42 & 56.06 & 71.07 & 79.64 & 72.84 & 73.38 & 27.42 & 68.02 & 54.59 \\
\cellcolor{lightblue}\textbf{+\method{} (Ours)} & \cellcolor{lightblue}\textbf{60.58} & \cellcolor{lightblue}\textbf{74.75} & \cellcolor{lightblue}\textbf{63.66} & \cellcolor{lightblue}\textbf{80.41} & \cellcolor{lightblue}\textbf{83.48} & \cellcolor{lightblue}\textbf{77.78} & \cellcolor{lightblue}\textbf{85.61} & \cellcolor{lightblue}\textbf{41.78} & \cellcolor{lightblue}\textbf{80.41} & \cellcolor{lightblue}\textbf{56.42} \\
\midrule
ORSTA\mbox{-}R1 & 45.43 & 51.49 & 51.31 & 66.06 & 71.04 & 70.37 & 67.67 & 33.17 & 67.65 & 57.57 \\
\cellcolor{lightblue}\textbf{+\method{} (Ours)} & \cellcolor{lightblue}\textbf{58.89} & \cellcolor{lightblue}\textbf{68.50} & \cellcolor{lightblue}\textbf{62.71} & \cellcolor{lightblue}\textbf{74.94} & \cellcolor{lightblue}\textbf{79.41} & \cellcolor{lightblue}\textbf{76.30} & \cellcolor{lightblue}\textbf{85.82} & \cellcolor{lightblue}\textbf{34.22} & \cellcolor{lightblue}\textbf{77.03} & \cellcolor{lightblue}\textbf{63.07} \\
\bottomrule
\end{tabular}}
\end{table*}

\begin{table*}[htbp]
\centering
\setlength{\tabcolsep}{4pt}
\renewcommand{\arraystretch}{1.12}
\caption{Accuracy (\%) on the remaining \textsc{VMCBench} datasets: TextVQA, InfoVQA, DocVQA, OCRVQA, VizWiz, ChartQA, GQA, A\mbox{-}OKVQA, OKVQA, VQAv2. Rows are paired as baseline vs.\ +\method{} for each model.}
\label{tab:vmc-split-b}
\resizebox{\linewidth}{!}{
\begin{tabular}{lccccccccccc}
\toprule
\textbf{Model} & \textbf{TextVQA} & \textbf{InfoVQA} & \textbf{DocVQA} & \textbf{OCRVQA} & \textbf{VizWiz} & \textbf{ChartQA} & \textbf{GQA} & \textbf{A\mbox{-}OKVQA} & \textbf{OKVQA} & \textbf{VQAv2} \\
\midrule
R1\mbox{-}OneVision & 64.72 & 56.36 & 58.56 & 66.06 & 64.18 & 56.65 & 61.86 & 56.71 & 63.95 & 66.44 \\
\cellcolor{lightblue}\textbf{+\method{}} & \cellcolor{lightblue}\textbf{90.56} & \cellcolor{lightblue}\textbf{77.14} & \cellcolor{lightblue}\textbf{94.44} & \cellcolor{lightblue}\textbf{89.38} & \cellcolor{lightblue}\textbf{84.42} & \cellcolor{lightblue}\textbf{82.11} & \cellcolor{lightblue}\textbf{80.93} & \cellcolor{lightblue}\textbf{80.24} & \cellcolor{lightblue}\textbf{81.98} & \cellcolor{lightblue}\textbf{84.03} \\
\midrule
Ocean\mbox{-}R1 & 91.01 & 71.19 & 92.00 & 87.31 & 86.86 & 75.92 & 80.68 & 81.65 & 86.67 & 83.56 \\
\cellcolor{lightblue}\textbf{+\method{}} & \cellcolor{lightblue}\textbf{97.75} & \cellcolor{lightblue}\textbf{82.16} & \cellcolor{lightblue}\textbf{100.00} & \cellcolor{lightblue}\textbf{96.63} & \cellcolor{lightblue}\textbf{90.91} & \cellcolor{lightblue}\textbf{84.63} & \cellcolor{lightblue}\textbf{88.75} & \cellcolor{lightblue}\textbf{89.41} & \cellcolor{lightblue}\textbf{92.35} & \cellcolor{lightblue}\textbf{88.89} \\
\midrule
MM\mbox{-}Eureka & 88.54 & 73.73 & 88.42 & 91.71 & 83.09 & 79.36 & 82.40 & 79.06 & 85.68 & 83.56 \\
\cellcolor{lightblue}\textbf{+\method{}} & \cellcolor{lightblue}\textbf{94.83} & \cellcolor{lightblue}\textbf{82.75} & \cellcolor{lightblue}\textbf{96.88} & \cellcolor{lightblue}\textbf{93.26} & \cellcolor{lightblue}\textbf{90.44} & \cellcolor{lightblue}\textbf{88.99} & \cellcolor{lightblue}\textbf{87.04} & \cellcolor{lightblue}\textbf{86.82} & \cellcolor{lightblue}\textbf{93.33} & \cellcolor{lightblue}\textbf{87.96} \\
\midrule
ORSTA\mbox{-}R1 & 83.37 & 70.43 & 82.20 & 86.79 & 82.99 & 76.15 & 72.86 & 72.94 & 75.80 & 81.02 \\
\cellcolor{lightblue}\textbf{+\method{}} & \cellcolor{lightblue}\textbf{95.28} & \cellcolor{lightblue}\textbf{86.71} & \cellcolor{lightblue}\textbf{97.97} & \cellcolor{lightblue}\textbf{93.52} & \cellcolor{lightblue}\textbf{89.18} & \cellcolor{lightblue}\textbf{86.70} & \cellcolor{lightblue}\textbf{81.66} & \cellcolor{lightblue}\textbf{80.71} & \cellcolor{lightblue}\textbf{88.89} & \cellcolor{lightblue}\textbf{87.27} \\
\bottomrule
\end{tabular}}
\end{table*}

\begin{table*}[htbp]
\centering
\setlength{\tabcolsep}{4pt}
\renewcommand{\arraystretch}{1.12}
\caption{\textsc{POPE}: accuracy (\%) and F1 by split (Adversarial / Popular / Random / Average).}
\resizebox{.7\linewidth}{!}{
\begin{tabular}{lcccccccc}
\toprule
\multirow{2}{*}{\textbf{Model}} & \multicolumn{2}{c}{\textbf{Adversarial}} & \multicolumn{2}{c}{\textbf{Popular}} & \multicolumn{2}{c}{\textbf{Random}} & \multicolumn{2}{c}{\textbf{Average}} \\
\cmidrule(lr){2-3}\cmidrule(lr){4-5}\cmidrule(lr){6-7}\cmidrule(lr){8-9}
 & \textbf{Acc} & \textbf{F1} & \textbf{Acc} & \textbf{F1} & \textbf{Acc} & \textbf{F1} & \textbf{Acc} & \textbf{F1} \\
\midrule
R1\mbox{-}OneVision & 69.20\% & 0.6629 & 70.33\% & 0.6711 & 71.12\% & 0.6769 & 70.22\% & 0.6703 \\
\cellcolor{lightblue}\textbf{+\method{}} & \cellcolor{lightblue}\textbf{80.90\%} & \cellcolor{lightblue}\textbf{0.7731} & \cellcolor{lightblue}\textbf{81.60\%} & \cellcolor{lightblue}\textbf{0.7825} & \cellcolor{lightblue}\textbf{82.43\%} & \cellcolor{lightblue}\textbf{0.7881} & \cellcolor{lightblue}\textbf{81.64\%} & \cellcolor{lightblue}\textbf{0.7805} \\
\midrule
Ocean\mbox{-}R1 & \textbf{85.53\%} & \textbf{0.8481} & \textbf{86.80\%} & \textbf{0.8592} & \textbf{87.97\%} & \textbf{0.8707} & \textbf{86.77\%} & \textbf{0.8593} \\
\cellcolor{lightblue}\textbf{+\method{}} & \cellcolor{lightblue}84.43\% & \cellcolor{lightblue}0.8271 & \cellcolor{lightblue}86.13\% & \cellcolor{lightblue}0.8458 & \cellcolor{lightblue}86.70\% & \cellcolor{lightblue}0.8496 & \cellcolor{lightblue}85.76\% & \cellcolor{lightblue}0.8408 \\
\midrule
MM\mbox{-}Eureka & 74.17\% & 0.7107 & 74.21\% & 0.7108 & 76.63\% & 0.7308 & 75.00\% & 0.7174 \\
\cellcolor{lightblue}\textbf{+\method{}} & \cellcolor{lightblue}\textbf{80.73\%} & \cellcolor{lightblue}\textbf{0.7726} & \cellcolor{lightblue}\textbf{81.17\%} & \cellcolor{lightblue}\textbf{0.7761} & \cellcolor{lightblue}\textbf{82.25\%} & \cellcolor{lightblue}\textbf{0.7860} & \cellcolor{lightblue}\textbf{81.38\%} & \cellcolor{lightblue}\textbf{0.7782} \\
\midrule
ORSTA\mbox{-}R1 & 68.31\% & 0.6445 & 72.00\% & 0.6754 & 73.77\% & 0.6866 & 71.36\% & 0.6688 \\
\cellcolor{lightblue}\textbf{+\method{}} & \cellcolor{lightblue}\textbf{81.86\%} & \cellcolor{lightblue}\textbf{0.7895} & \cellcolor{lightblue}\textbf{83.39\%} & \cellcolor{lightblue}\textbf{0.8061} & \cellcolor{lightblue}\textbf{83.67\%} & \cellcolor{lightblue}\textbf{0.8071} & \cellcolor{lightblue}\textbf{82.98\%} & \cellcolor{lightblue}\textbf{0.8009} \\
\bottomrule
\end{tabular}}
\end{table*}

\subsection{Transfer \method{} to Non-Reasoning MLLMs}
\label{sec:non-reasoning}
\begin{figure*}[htbp]
    \centering
    \includegraphics[width=.95\textwidth]{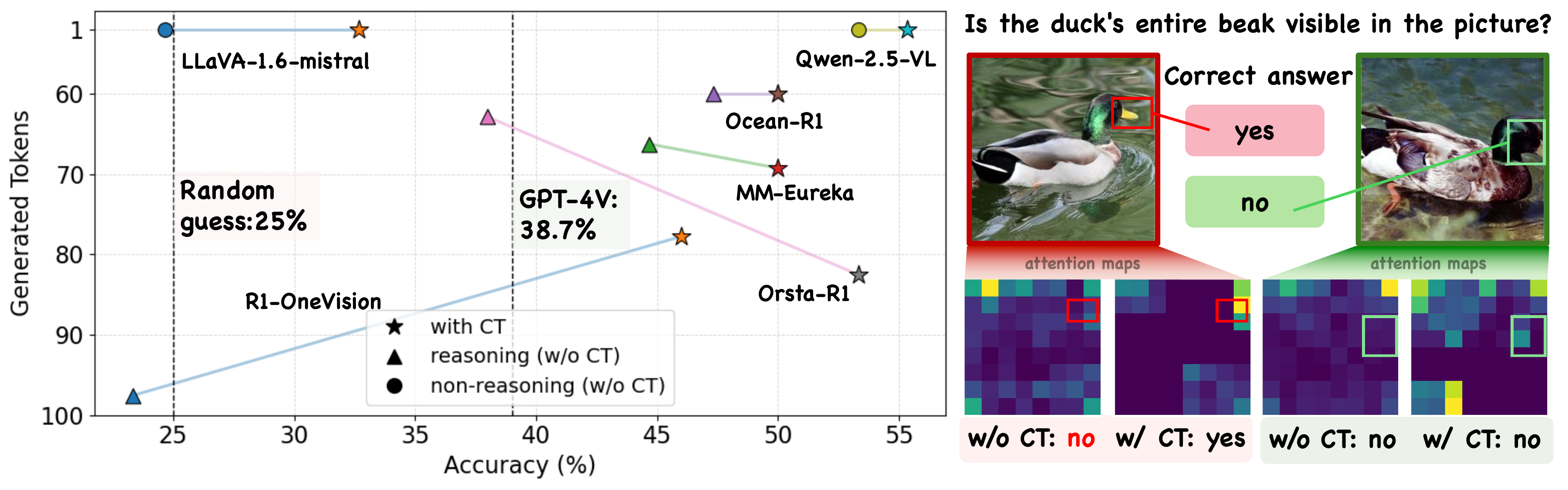}
    \caption{\textbf{Left:} Accuracy versus output–token length on \textsc{MMVP} across models; \textbf{Right:} attention allocation with and without \method{}.}
    \label{fig:rq4}
\end{figure*}

\textbf{Reasoning subset members.}
\emph{MMMU}: college\mbox{-}level, multi\mbox{-}discipline problems requiring deliberative reasoning \cite{yue2023mmmu}. 
\emph{MathVista}: mathematical reasoning in visual contexts (charts, plots, geometry, etc.) \cite{lu2023mathvista}. 
\emph{MathVision}: math problem solving on real\mbox{-}world photos and synthetic scenes \cite{wang2024mathvision}. 
\emph{MMStar}: broad, multi\mbox{-}skill evaluation of multimodal reasoning \cite{chen2024mmstar}.

\textbf{General subset members.}
\emph{VQAv2}: balanced open\mbox{-}ended VQA over everyday photos \cite{goyal2017making}. 
\emph{GQA}: compositional visual reasoning with scene\mbox{-}graph grounding \cite{hudson2019gqa}. 
\emph{VizWiz}: real\mbox{-}world VQA from blind photographers; noisy images and conversational questions \cite{gurari2018vizwiz}. 
\emph{OK\mbox{-}VQA}/\emph{A\mbox{-}OKVQA}: knowledge\mbox{-}based VQA requiring external/world knowledge \cite{marino2019okvqa,schwenk2022aokvqa}.

\textbf{OCR subset members.}
\emph{TextVQA}: reading text in the wild for VQA \cite{singh2019textvqa}. 
\emph{OCR\mbox{-}VQA}: VQA by reading textual content in images (book covers, storefronts, etc.) \cite{mishra2019ocrvqa}.

\textbf{Doc \& Chart subset members.}
\emph{DocVQA}: document understanding for VQA (forms, invoices, pages) \cite{mathew2021docvqa}. 
\emph{TableVQA\mbox{-}Bench}: multi\mbox{-}domain table QA benchmark for visual tables \cite{kim2024tablevqabench}. 
\emph{InfographicVQA}: QA over complex infographics requiring reading and reasoning \cite{mathew2022infographicvqa}. 
\emph{AI2D}: diagram understanding and diagram\mbox{-}based QA \cite{kembhavi2016diagram}. 
\emph{ChartQA}: QA over charts requiring numerical and logical reasoning \cite{masry2022chartqa}.
\paragraph{Additional sets in \textsc{VMCBench}.}
\emph{MM\mbox{-}Vet}: comprehensive capability evaluation of MLLMs, frequently used as a generalization stress test \cite{yu2023mmvet}. 
\emph{SEED\mbox{-}Bench}: multi\mbox{-}domain, fine\mbox{-}grained evaluation for perception and reasoning \cite{li2024seedbench}. 
\emph{RealWorldQA}: real\mbox{-}world spatial understanding; images largely from vehicles with verifiable Q\&A \cite{xai2024realworldqa}.

\paragraph{Thinking length.}{ We measure the average output token length on the ``general” and ``reasoning” subsets of VMC-Bench. With \method{}, the average output length on the reasoning subset increases from 197.22 to 302.88 tokens (a 53.57\% increase), while on the general subset it only increases from 134.43 to 136.10 tokens (a 1.24\% increase). This further supports that \method{} reduces the perceptual burden and lets the model use more capacity for reasoning.}

\paragraph{Transfer to Non-Reasoning MLLMs.}
To verify the transferability of \method{}, we evaluate \textsc{MMVP} on \emph{non-reasoning} MLLMs. Concretely, we adopt the \textit{offline} setting: attention-guided crops produced by \texttt{Ocean-R1} are fed to \texttt{Qwen2.5-VL} and \texttt{LLaVA-1.6-Mistral}. Results are summarized in Fig.~\ref{fig:rq4}. Our findings are:
\textsc{MMVP} is designed to probe visual defects in multimodal models. As shown in Fig.~\ref{fig:rq4}, \method{} consistently improves performance; in particular, it moves several models—\texttt{LLaVA-1.6}, \texttt{R1-OneVision}, and \texttt{ORSTA-R1}—from near random guessing toward the \texttt{GPT-4V} level. In the qualitative examples of Fig.~\ref{fig:rq4}, without \method{} the models barely attend to the key region of the question, whereas with \method{} they shift focus to that region.

\subsection{Results on RWQA and 3DSRBench}
\label{subsec:rwqa-3dsr}
{
We evaluate \method{} on RWQA \cite{xai2024realworldqa} and 3DSRBench \cite{ma20253dsrbench} using three base models. As shown in Table~\ref{tab:rwqa-3dsr}, \method{} consistently improves accuracy on both datasets for all backbones.
}
\begin{table*}[htbp]
    \centering
    \caption{{Results on RWQA and 3DSRBench with and without \method{}.}}
    \label{tab:rwqa-3dsr}
    \begin{tabular}{lccccc}
        \toprule
        Dataset   & Setting       & Ocean-R1 & ORSTA  & MM-Eureka \\
        \midrule
        RWQA      & w/o \method{}   & 63.27\%  & 65.23\% & 60.39\% \\
        RWQA      & with \method{}  & 70.33\%  & 67.84\% & 65.49\% \\
        3DSRBench & w/o \method{}   & 37.66\%  & 43.44\% & 56.25\% \\
        3DSRBench & with \method{}  & 40.06\%  & 45.14\% & 59.24\% \\
        \bottomrule
    \end{tabular}
\end{table*}

\subsection{Ablation Result Discussions}
\label{sec:app:dis}
In this section, we discuss the ablation results in detail.

\begin{figure*}[htbp]
    \centering
    \includegraphics[width=1\linewidth]{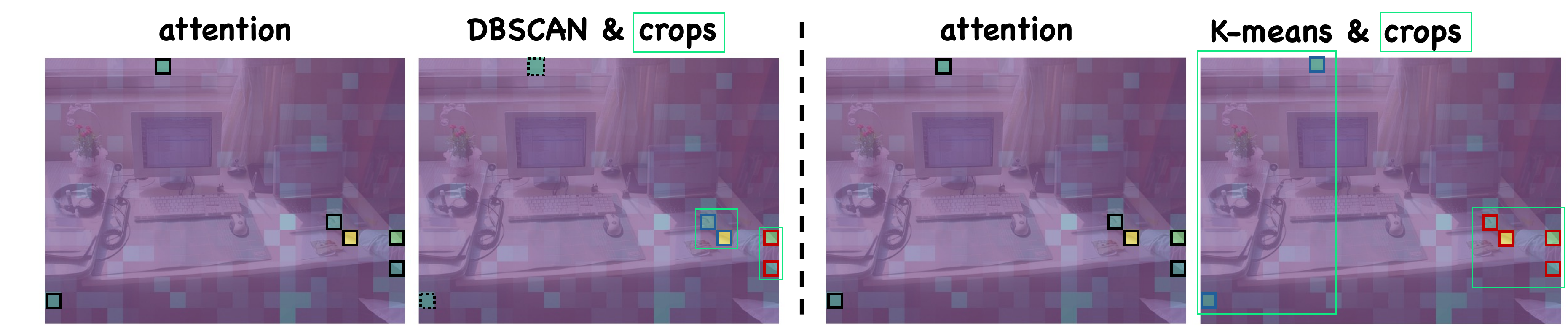}
    \caption{\textbf{Case study on clustering methods.} Each subfigure has two parts. \emph{Left:} \method{} attention, i.e., a visualization of the attention map, where the \method{}-selected visual patches \(\mathcal{V}_k^{\star}\) are outlined with black boxes. \emph{Right:} clustering results from different methods. Because K\mbox{-}means is sensitive to noise, the resulting crops include large amounts of question-irrelevant content induced by only two noisy points, whereas DBSCAN avoids this issue.}
    \vspace{-1em}
    \label{fig:abla_clus}
\end{figure*}

\begin{figure*}[htbp]
    \centering
    \includegraphics[width=.98\linewidth]{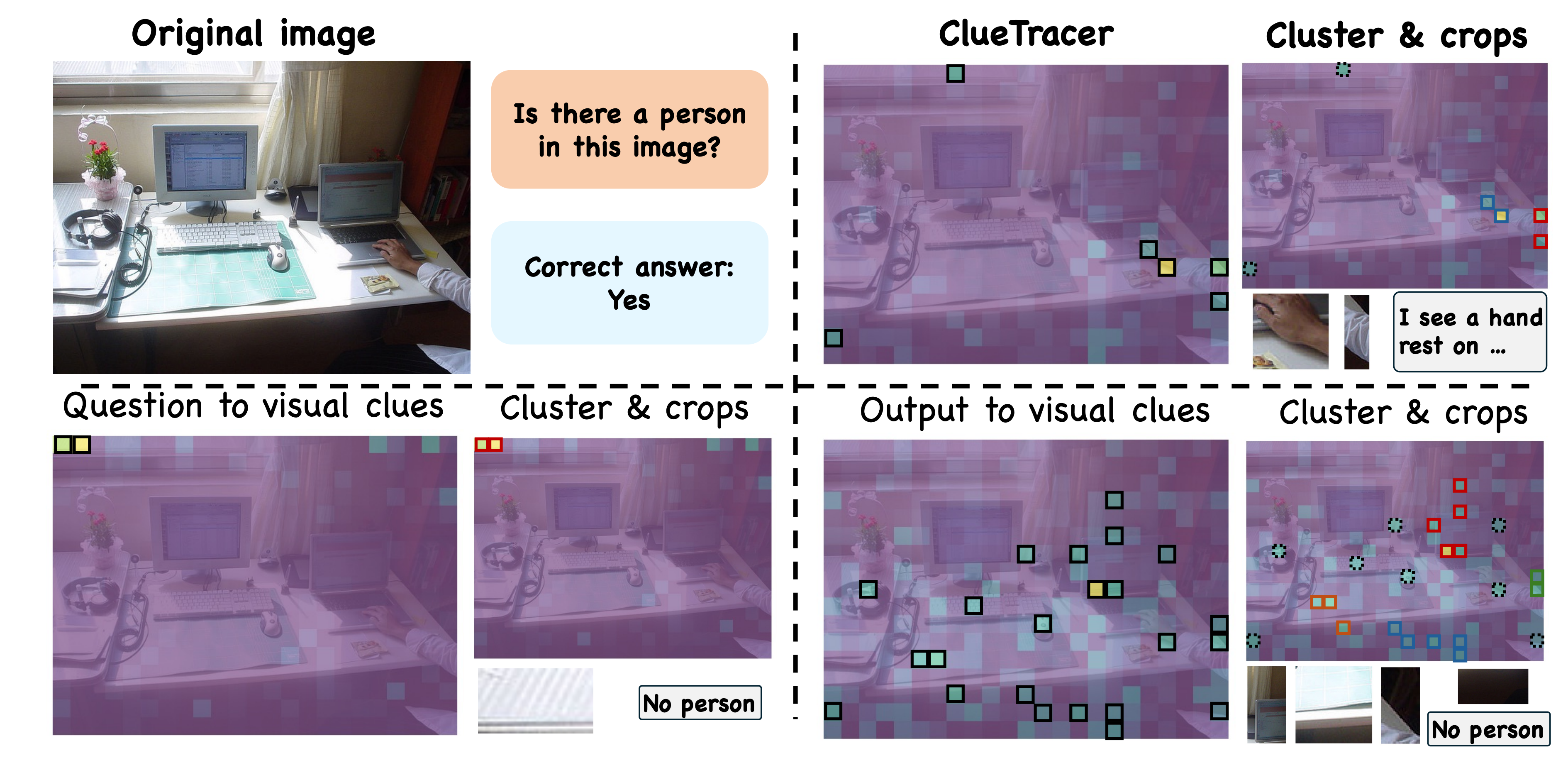}
    \caption{\textbf{Case study on attention choices.} Attention identifies question-relevant tokens and correctly crops the hand, thereby guiding the model to the correct answer. \emph{Question to visual} denotes cross-attention between the input question and input visual tokens; because the model is not trained for this alignment, the signal is weak. \emph{Output to visual} directly leverages generation-side attention, which is mainstream in MLLMs; however, in MLRMs, reasoning drift introduces numerous distractors.}
    \vspace{-1em}
    \label{fig:abla_attn}
\end{figure*}

\paragraph{Clustering methods.}
We compare \textsc{K\mbox{-}means} and \textsc{DBSCAN} at the patch clustering step. \textsc{K\mbox{-}means} is a partition-based method, forcing every patch to be assigned to some cluster; this makes it sensitive to noise. In contrast, \textsc{DBSCAN} is density-based and permits points outside high-density regions to remain as noise. Practically, \textsc{K\mbox{-}means} tends to absorb noisy patches into clusters, yielding overly large crops, whereas \textsc{DBSCAN} better isolates compact evidence regions. An example is provided in Fig.~\ref{fig:abla_clus}: \textsc{K\mbox{-}means} nearly crops the entire image, while \textsc{DBSCAN} focuses on the key area.
\paragraph{Attention choices.}
We compare \method{} with two baselines: \emph{output$\rightarrow$visual} attention (common in generative pipelines) and \emph{question$\rightarrow$visual} attention (input–input cross-attention, common in discriminative settings); see Fig.~\ref{fig:abla_attn}. Due to \emph{reasoning drift}, output$\rightarrow$visual alone can be distracted by complex scenes and fail to attend to the question-relevant region. Conversely, question$\rightarrow$visual underutilizes the model’s training signal, which is primarily aligned to fitting the output sequence, leading to a mismatch and degraded localization.

\subsection{Case Study}
\begin{figure*}[htbp]
    \centering
    \includegraphics[width=1\textwidth]{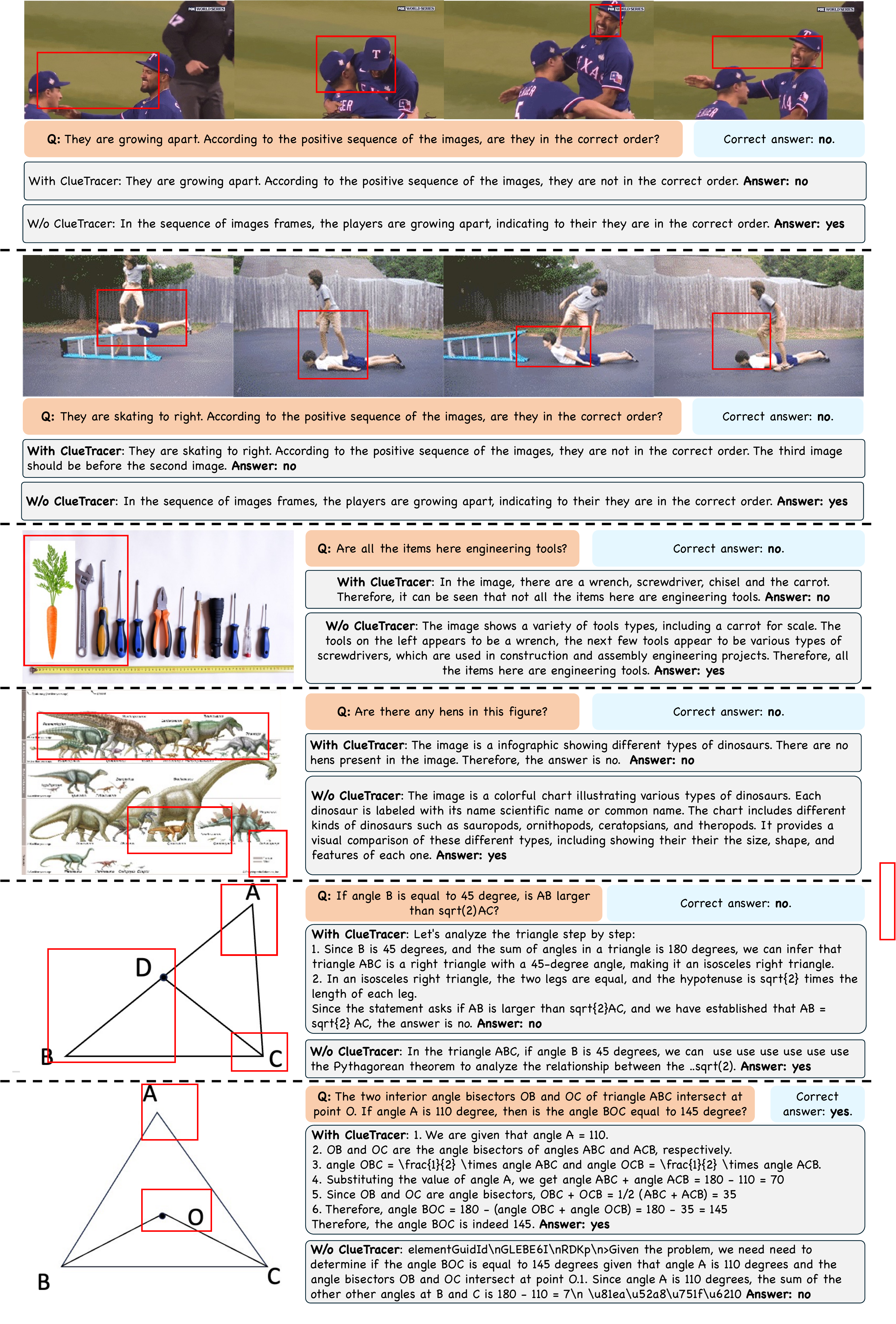}
    \caption{\textbf{Case study}~(Part One). The visual crops selected by \method{} are highlighted with red bounding boxes in the figure.}
    \vspace{-1em}
    \label{fig:appendix_case_01}
\end{figure*}

\begin{figure*}[htbp]
    \centering
    \includegraphics[width=1\textwidth]{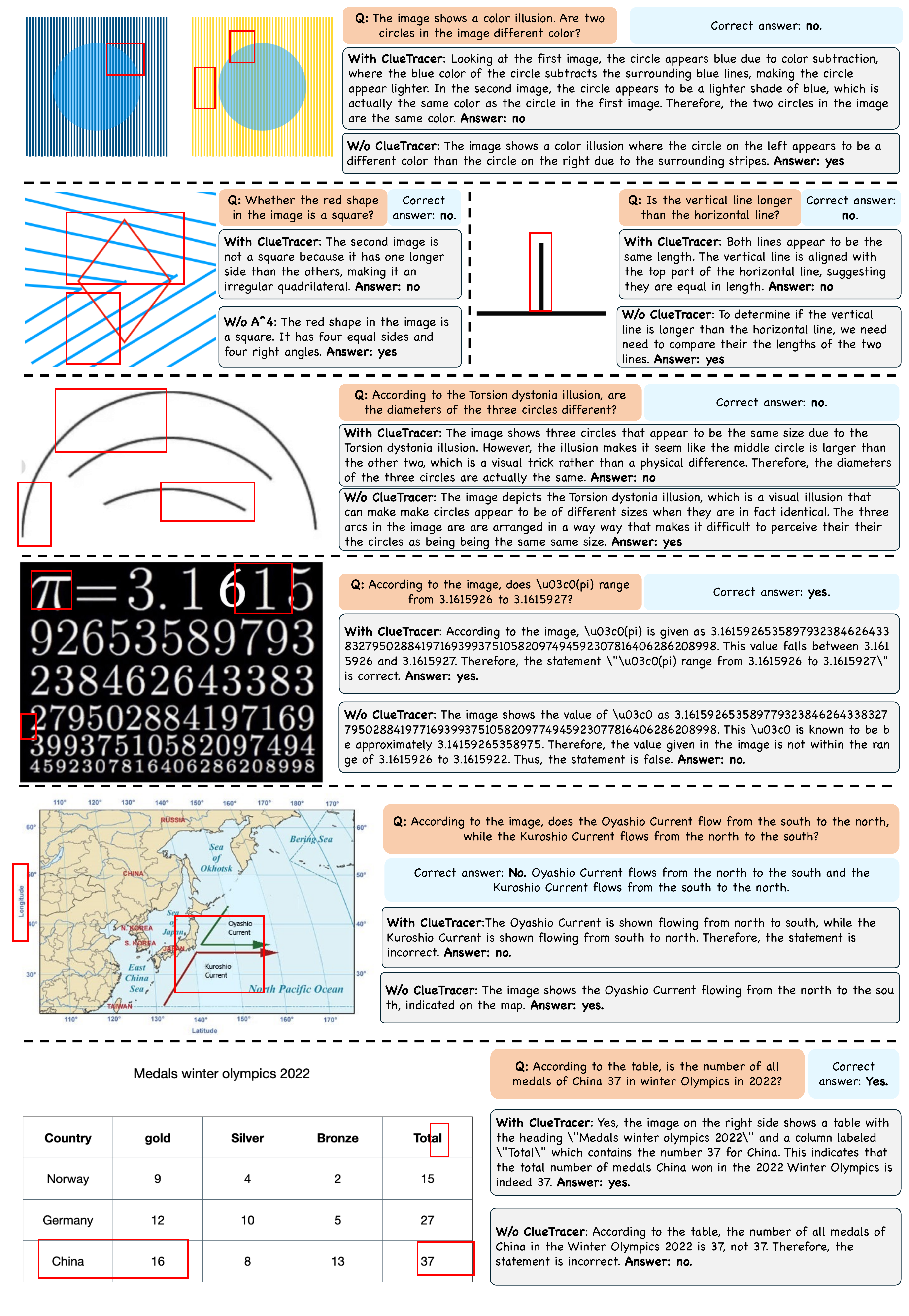}
    \caption{\textbf{Case study}~(Part Two). The visual crops selected by \method{} are highlighted with red bounding boxes in the figure.}
    \vspace{-1em}
    \label{fig:appendix_case_02}
\end{figure*}

\begin{figure*}[htbp]
    \centering
    \includegraphics[width=0.85\textwidth]{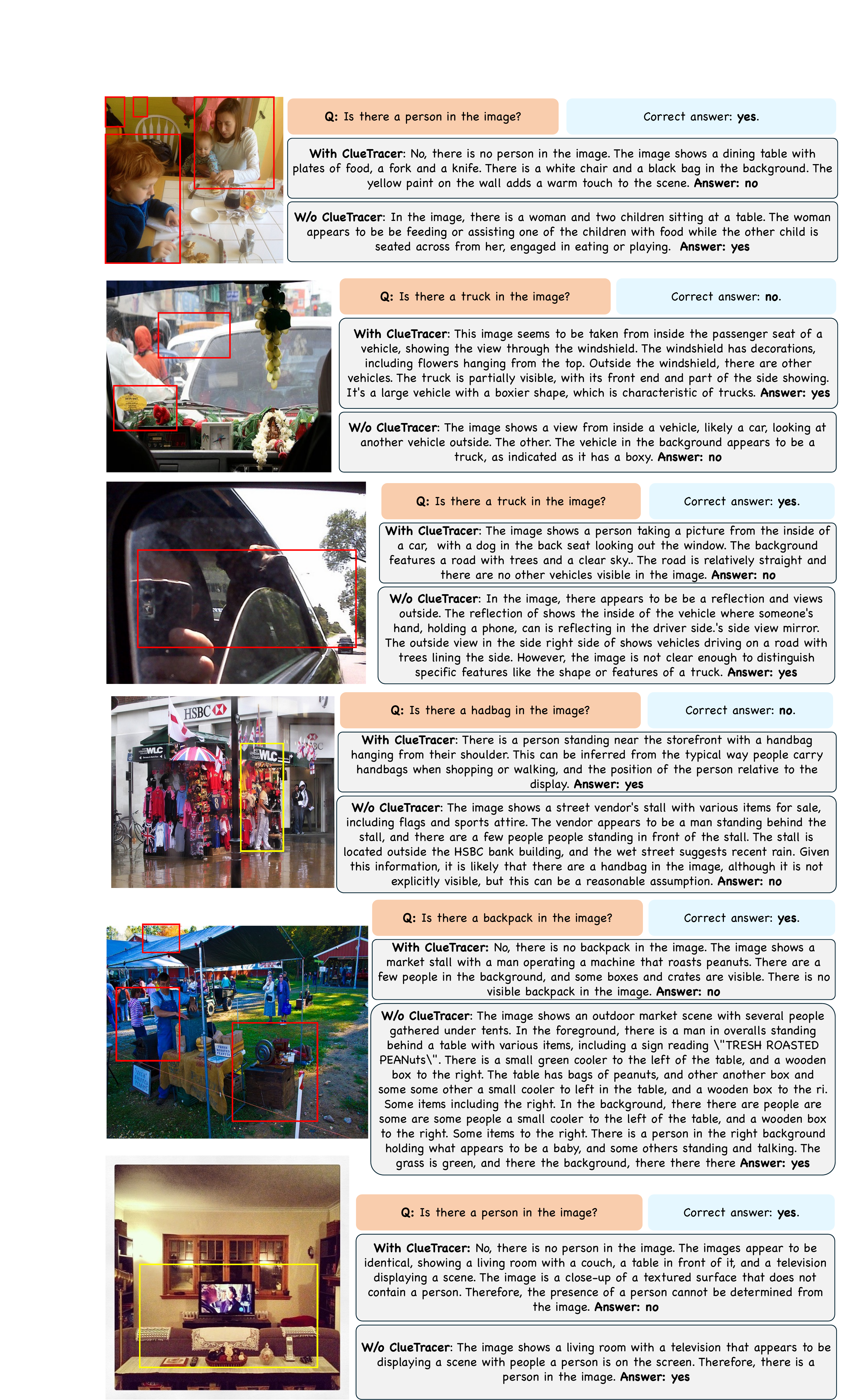}
    \caption{{\textbf{Failure case study}~(Part One). The visual crops selected by \method{} are highlighted with red and yellow bounding boxes. We observe that, in most cases, \method{} crops the correct regions, but it fails to do so in the third example.}}
    \vspace{-1em}
    \label{fig:appendix_case_03}
\end{figure*}

\begin{figure*}[htbp]
    \centering
    \includegraphics[width=1\textwidth]{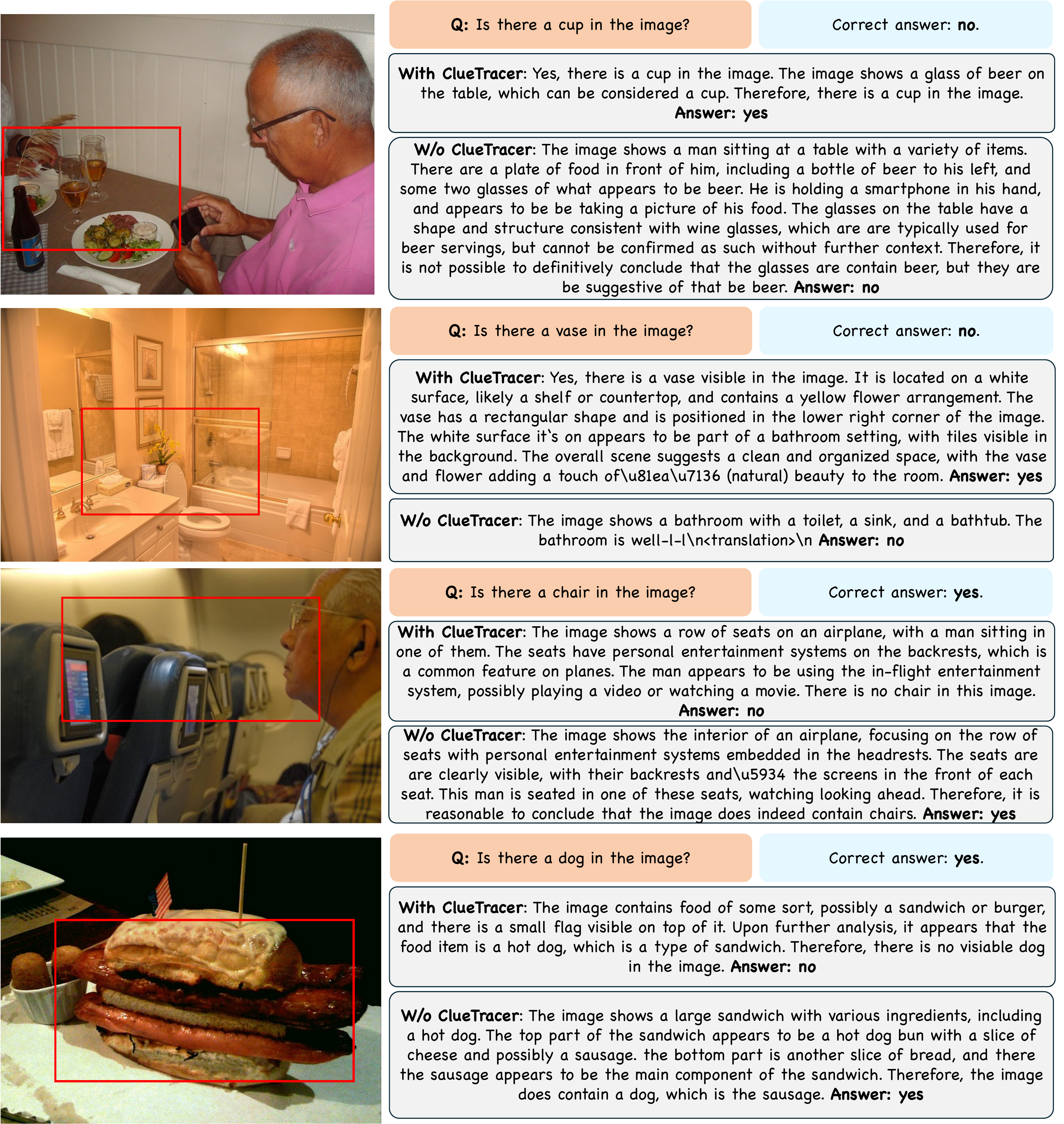}
    \caption{{\textbf{Failure case study}~(Part Two). The visual crops selected by \method{} are highlighted with red/yellow boxes. We observe that these failures are driven by object-level semantics: in these examples, the model recognizes semantically similar but incorrect concepts (e.g., \textit{glass} vs.\ \textit{cup}, \textit{seat} vs.\ \textit{chair}).}}
    \vspace{-1em}
    \label{fig:appendix_case_04}
\end{figure*}

We present side-by-side comparisons across multiple reasoning models \emph{with} and \emph{without} \method{}. As shown in Fig.~\ref{fig:appendix_case_01} and Fig.~\ref{fig:appendix_case_02}, red boxes denote the visual crops. \method{} consistently localizes task-relevant regions and produces more accurate answers than the baseline. We also report failure cases in Fig.~\ref{fig:appendix_case_03} and Fig.~\ref{fig:appendix_case_04}, where Fig.~\ref{fig:appendix_case_03} illustrates failures induced by visual semantics and Fig.~\ref{fig:appendix_case_04} illustrates failures induced by query semantics.
\label{sec:app:case}

\section{q2o and v2o Heatmaps}
\label{sec:app:moti}
\vspace{-0.4em}
\textbf{Setup and notation.}
We adopt the standard Transformer attention view for decoder-style MLRMs~\cite{vaswani2017attention}: at output step $t$, the model attends over the input-side question tokens $\mathbf{X}_q = \{x_{q,i}\}_{i=1}^{N_q}$ and visual tokens $\mathbf{X}_v = \{x_{v,j}\}_{j=1}^{N_v}$, producing head-aggregated, key-normalized weights $A^{q\to o}_{t,i}\in[0,1]$ and $A^{v\to o}_{t,j}\in[0,1]$ with $\sum_i A^{q\to o}_{t,i}=1$ and $\sum_j A^{v\to o}_{t,j}=1$. Unless otherwise noted we average across layers/heads (a common practice in attention-based attribution \emph{post hoc} analyses). COCO categories and bounding boxes $(x,y,w,h)$ are denoted by \texttt{cat} and \texttt{bbox} and follow the dataset’s instance annotations~\cite{lin2014microsoft}.

\textbf{q2o heatmap (question $\rightarrow$ output).}
Given an object string \texttt{cat}, we first detect \emph{object mentions} in the generated output $\{y_t\}_{t=1}^{T}$ by letters-only matching (case-insensitive) with a leniency margin (token length $\le |\texttt{cat}|{+}2$) and POS tag \textsc{NOUN}. Let the set of indices of these mentions be $\mathcal{T}_{\mathrm{obj}}$. For every output token $t$ with POS \textsc{NOUN}, define its distance to the nearest object mention $d(t)=\min_{u\in\mathcal{T}_{\mathrm{obj}}}|t-u|$. We form seven noun buckets $B_d=\{t:\mathrm{POS}(t)=\textsc{NOUN},\,d(t)=d\}$ for $d\in\{0,\dots,6\}$, and three POS buckets over all outputs $B_{\mathrm{ADJ}},B_{\mathrm{VERB}},B_{\mathrm{OTHER}}$. The $N_q\times 10$ matrix $M$ used for the q2o heatmap is
\[
\begin{aligned}
M_{i,k}
&= \tfrac{1}{|B_k|}\sum_{t\in B_k}A^{q\to o}_{t,i} \\
&\quad \text{for } k\in\{0,\dots,6,\mathrm{ADJ},\mathrm{VERB},\mathrm{OTHER}\}.
\end{aligned}
\]
with empty buckets contributing zeros. We visualize $M$ as a diverging heatmap (high\,$\rightarrow$\,red, low\,$\rightarrow$\,blue) with smooth interpolation; the x-axis shows different buckets, and the y-axis lists question tokens.

\textbf{v2o heatmap (visual $\rightarrow$ output around a bbox).}
For each COCO bounding box \texttt{bbox} of category \texttt{cat} present in the output, we locate the nearest occurrence index $t_{\texttt{cat}}$ of \texttt{cat} (letters-only matching; multiword matched as a contiguous phrase). We extract a temporal window $W=\{t_{\texttt{cat}}-10,\dots,t_{\texttt{cat}}+10\}\cap[1,T]$ (21 positions) and the corresponding rows of the visual-to-output attention, giving $A^{v\to o}_{W,:}\in\mathbb{R}^{21\times N_v}$. Using the patch grid $(t_h,t_w)$ from the resized image, we map \texttt{bbox} to its covered patch set $P_{\texttt{bbox}}\subseteq\{1,\dots,N_v\}$ by IoU$>0.5$ between \texttt{bbox} and each patch cell (pixel-aligned rectangles). Let $r_{\min}\!:\!r_{\max}$ and $c_{\min}\!:\!c_{\max}$ be the row/column spans of $P_{\texttt{bbox}}$ on the $(t_h{\times}t_w)$ grid. For lateral offsets $s\in\{-5,\dots,5\}$ we define expanded regions
\[
\resizebox{\columnwidth}{!}{$
\begin{aligned}
P_{\texttt{bbox},s}
&= \{(r,c): r\in I_r,\ c\in I_{c,s}\}
   \cap [0,t_h\!-\!1]\times[0,t_w\!-\!1],\\
I_r&=[r_{\min},r_{\max}],\\
I_{c,s}&=[c_{\min}+\min(0,s),\,c_{\max}+\max(0,s)].
\end{aligned}
$}
\]
flattened to patch indices. The $21\times 11$ matrix $V$ for \texttt{bbox} is
\[
V_{r,\,s+5}=\tfrac{1}{|P_{\texttt{bbox},s}|}\sum_{j\in P_{\texttt{bbox},s}} A^{v\to o}_{t_r,\,j},\qquad t_r\in W,
\]
summarizing how attention over visual patches varies near the bounding box versus its lateral background. We render $V$ with the same diverging colormap and smooth interpolation; the y-axis uses three ticks (\texttt{others}–\{\texttt{cat}\}–\texttt{others}) centered at $t_{\texttt{cat}}$, and the x-axis uses three ticks (\texttt{background}–\texttt{bbox patches}–\texttt{background}) centered at $s{=}0$.

\section{Sensitivity, scalability, and deployment cost analysis}
\label{subsec:more-sensitivity}
{
We further analyze the sensitivity of \method{} to the thresholds $\tau_q$, $\tau_o$, $\tau_v$ and the clustering parameter dbscan-$\epsilon$. As shown in Tables~\ref{tab:sens-tauq}--\ref{tab:sens-dbscan}. We also summarize latency and throughput in Table~\ref{tab:latency-throughput}, and report layer-wise AttnRecall on larger 32B models and LLaVA architecture \cite{liu2023improved} in Table~\ref{tab:attnrecall-32b} and Table \ref{tab:attnrecall-llava}.}
\begin{table*}[htbp]
    \centering
    \caption{{Sensitivity of accuracy to $\tau_q$ on HallusionBench and POPE.}}
    \label{tab:sens-tauq}
    \begin{tabular}{lccccc}
        \toprule
        Method & 0.5 & 1 & 1.5 & 2 & 3 \\
        \midrule
        HallusionBench + MM-Eureka & 63.51 & 65.10 & 63.51 & 63.15 & 61.65 \\
        POPE + MM-Eureka           & 79.86 & 81.38 & 80.32 & 79.13 & 80.92 \\
        POPE + Orsta               & 83.32 & 83.67 & 85.01 & 83.12 & 82.91 \\
        \bottomrule
    \end{tabular}
\end{table*}

\begin{table*}[htbp]
    \centering
    \caption{{Sensitivity of accuracy to $\tau_o$ on HallusionBench and POPE.}}
    \label{tab:sens-tauo}
    \begin{tabular}{lccccc}
        \toprule
        Method & 0.5 & 1 & 1.5 & 2 & 3 \\
        \midrule
        HallusionBench + MM-Eureka & 66.43 & 65.10 & 63.86 & 63.42 & 63.77 \\
        POPE + MM-Eureka           & 82.32 & 81.38 & 79.22 & 79.62 & 80.62 \\
        POPE + Orsta               & 84.42 & 83.67 & 85.02 & 83.12 & 79.02 \\
        \bottomrule
    \end{tabular}
\end{table*}
\begin{table*}[htbp]
    \centering
    \caption{{Sensitivity of accuracy to $\tau_v$ on HallusionBench and POPE.}}
    \label{tab:sens-tauv}
    \begin{tabular}{lccccc}
        \toprule
        Method & 0.5 & 1 & 1.5 & 2 & 3 \\
        \midrule
        HallusionBench + MM-Eureka & 64.22 & 65.46 & 65.10 & 63.95 & 63.45    \\
        POPE + MM-Eureka           & 82.92 & 82.41 & 81.38 & 80.92 & 79.12 \\
        POPE + Orsta               & 81.82 & 83.22 & 83.67 & 85.81 & 83.22 \\
        \bottomrule
    \end{tabular}
\end{table*}
\begin{table*}[htbp]
    \centering
    \caption{{Sensitivity of accuracy to DBSCAN-$\epsilon$ on HallusionBench and POPE.}}
    \label{tab:sens-dbscan}
    \begin{tabular}{lcccccc}
        \toprule
        Method & 1.1 & 1.9 & 2.1 & 2.9 & 3.1 & 3.9 \\
        \midrule
        HallusionBench + MM-Eureka & 63.31 & 64.57 & 64.48 & 65.10 & 65.20 & 65.37 \\
        POPE + MM-Eureka           & 78.52 & 79.62 & 80.62 & 81.38 & 81.32 & 81.22 \\
        POPE + Orsta               & 84.52 & 84.62 & 85.41 & 83.67 & 84.52 & 83.02 \\
        \bottomrule
    \end{tabular}
\end{table*}

\paragraph{Sensitivity Analysis. }
{As summarized in Tables~\ref{tab:sens-tauq}--\ref{tab:sens-dbscan}, the accuracies vary only mildly across a wide range of $\tau_q$, $\tau_o$, $\tau_v$, and dbscan-$\epsilon$, indicating that \method{} is not overly sensitive to these hyperparameters. In all main experiments, we therefore fix $\tau_q = 1$, $\tau_o = 1$, $\tau_v = 1.5$, and \text{dbscan-}$\epsilon$ = 2.9.}
\paragraph{Deployment Cost Analysis. } {We conduct a deployment cost analysis of \method{} and report end-to-end costs under different Stage 1 output lengths. All experiments are run with a 7B model on a single NVIDIA GeForce RTX 3090 (20G). As shown in Table~\ref{tab:latency-throughput}, longer Stage 1 sequences provide richer priors to Stage 2 but also incur higher inference cost, which is consistent with our expectation.}
\begin{table*}[htbp]
    \centering
    \caption{{Latency and throughput under different online and offline settings.}}
    \label{tab:latency-throughput}
    \resizebox{.8\textwidth}{!}{
    \begin{tabular}{lcccccc}
        \toprule
        Mode / sequence length       & offline/-- &  1 & 20 & 50 & 128 & 256 \\
        \midrule
        Latency (hours / 1000 items) & 0.91         & 0.97       & 1.08        & 1.36        & 1.62         & 1.79 \\
        Throughput (tokens/s)        & 39.80        & 35.52      & 31.24       & 27.71       & 20.91        & 18.08 \\
        \bottomrule
    \end{tabular}
    }
\end{table*}

\paragraph{Scalability Analysis. }
{As shown in table \ref{tab:attnrecall-32b} and \ref{tab:attnrecall-llava}, on both 32B variants and models with different underlying architectures, the layer-wise \metric{} exhibits the same rise then fall pattern as in 7B models, and the peak perceptual layers still lie in the later part of the network. For both Orsta-32B and MM Eureka-32B, the global maximum of \metric{} appears around layer 40 out of 64, which corresponds to roughly 70\% of the total depth. This indicates that our perception layer finding scales to larger parameter regimes.}

\section{Potential Risks}
This work aims to reduce hallucinations in multimodal reasoning models, but it may also introduce potential risks if over-relied upon in high-stakes applications. Since \method{} improves visual grounding by tracing model attention, its outputs may give users a stronger sense of reliability even when the underlying model still makes incorrect judgments.

\begin{table*}[htbp]
    \centering
    \caption{{Layer-wise \metric{} on 32B models.}}
    \label{tab:attnrecall-32b}
    \resizebox{\textwidth}{!}{
    \begin{tabular}{lcccccccccc}
        \toprule
        Model         & 0      & 12     & 24     & 30     & 36     & \textbf{40} & 42     & 48     & 54     & 63     \\
        \midrule
        MM-Eureka-32B & 0.3060 & 0.3377 & 0.4978 & 0.5358 & 0.5781 & \textbf{0.6271} & 0.5474 & 0.4854 & 0.3992 & 0.3099 \\
        Orsta-32B     & 0.3165 & 0.3397 & 0.4922 & 0.5381 & 0.5669 & \textbf{0.6350} & 0.5596 & 0.4961 & 0.4059 & 0.3172 \\
        \bottomrule
    \end{tabular}
    }
\end{table*}

\begin{table*}[htbp]
    \centering
    \caption{{Layer-wise \metric{} on LLaVA-1.5-7B and LLaVA-Mistral-7B.}}
    \label{tab:attnrecall-llava}
    \resizebox{\textwidth}{!}{
    \begin{tabular}{lcccccccc}
        \toprule
        model            & Arch       & 0       & 6       & 12      & 17      & 18      & 24      & 31      \\
        \midrule
        LLaVA-1.5-7B     & LLaVA-1.5  & 33.81\% & 38.00\% & 43.04\% & \textbf{50.59\%} & 48.29\% & 47.03\% & 39.02\% \\
        LLaVA-Mistral-7B & LLaVA-NeXT &  10.37\%  & 23.95\%  & 46.00\% & 49.04\%  & \textbf{49.72\%} & 41.10\%   & 35.95\% \\
        \bottomrule
    \end{tabular}
    }
\end{table*}

\section{Discussion with Prior Attention-based Studies}
\subsection{Attention Applications in VLMs}
Prior works have explored visual attention in vision-language models for model interpretation, visual grounding, and attention analysis~\cite{10.3389/frai.2021.767971,cao-etal-2022-understanding}. For example, visual attention was used to inspect image-captioning models by revealing attended image regions during generation~\cite{xu2015show}, while later studies analyzed textual and cross-modal attention to understand how multimodal transformers ground entities and relations~\cite{ilinykh2022attention}. Related studies in referring expression generation further investigated how models allocate attention to target objects, spatial information, and contextual regions during visual language generation~\cite{zarrie2023keeping,merlo2026cooco}. 

Different from these studies, which mainly analyze attention patterns in conventional vision-language generation or object-level grounding, our work focuses on multimodal reasoning models and studies how intermediate outputs mediate visual evidence retrieval. Specifically, we introduce a question$\rightarrow$output$\rightarrow$vision tracing mechanism to identify task-critical clues during long-chain reasoning, enabling the characterization and mitigation of reasoning drift.

\subsection{Compatibility with Gradient-based Attribution}
\label{sec:app:gradcam}

Although \method{} uses native attention for efficient online inference, its query$\rightarrow$output$\rightarrow$vision evidence-tracing formulation is not restricted to attention-based attribution and can be integrated with gradient-based methods such as Grad-CAM~\cite{selvaraju2017gradcam}. Existing Grad-CAM applications to LVLMs typically attribute the final answer token to visual regions~\cite{zhang-etal-2025-redundancy}. This is suitable for short-form generation, but in long-chain reasoning, decisive visual evidence may be retrieved earlier in the reasoning trajectory, while the final answer can depend primarily on previously generated text. We therefore extend Grad-CAM with our clue-tracing strategy: instead of attributing only the final answer token, we identify question-relevant reasoning tokens and aggregate their Grad-CAM maps. We evaluate localization using the same bounding-box recall metric.

\begin{table}[t]
\centering
\small
\begin{tabular}{lcc}
\toprule
Method & Ocean-R1 & MM-Eureka \\
\midrule
Standard Grad-CAM & 0.267 & 0.274 \\
\method{} + Grad-CAM & \textbf{0.377} & \textbf{0.390} \\
\bottomrule
\end{tabular}
\caption{Compatibility of \method{} with Grad-CAM. Incorporating question-relevant reasoning tokens substantially improves visual clue localization over standard final-token attribution.}
\label{tab:gradcam}
\end{table}

These results further support that the core contribution of \method{} lies in the output-mediated evidence-tracing formulation, which can be instantiated with different attribution mechanisms.